\documentclass{article}

\usepackage[preprint]{neurips_2026}

\usepackage[utf8]{inputenc}
\usepackage[T1]{fontenc}
\usepackage{hyperref}
\usepackage{url}
\usepackage{booktabs}
\usepackage{graphicx}
\usepackage{adjustbox}
\usepackage{wrapfig}
\usepackage{multirow}
\usepackage{amsfonts}
\usepackage{amsmath}
\usepackage{amssymb}
\usepackage{amsthm}
\usepackage{algorithm}
\usepackage{algpseudocode}
\usepackage{nicefrac}
\usepackage{microtype}
\usepackage{xcolor}
\usepackage{tikz}
\usetikzlibrary{arrows.meta,positioning,fit,backgrounds,calc}

\title{Markovian Dynamics Enforcer: Feasibility Preserving Correction on Learned Dynamics Manifolds}

\author{%
  Kevin Yu$^{1}$\thanks{Correspondence to: \texttt{k.qu23@imperial.ac.uk}} \quad
  Tao Guo$^{2}$ \quad
  Constantinos Antoniou$^{2}$ \quad
  Panagiotis Angeloudis$^{1}$ \\
  $^{1}$Imperial College London, UK \quad
  $^{2}$Technical University of Munich, Germany \\
}

\begin{document}

\maketitle

\begin{abstract}
Neural trajectory predictors can reach low prediction error while violating dynamics, actuator limits, or state constraints, especially when controls are unobserved and dynamics are partially specified. We introduce the Markovian Dynamics Enforcer (MaDE)\footnote[1]{Code available at \url{https://github.com/tsl-imperial/MaDE}}, a time-invariant post-hoc operator mapping state-transition proposals onto a learned feasible dynamics manifold, trained on feasible states without ground-truth controls. For each transition it infers a control and recomputes the state through a completion model of known physics plus a learned residual. It then corrects that control by gradient-based inequality reduction, so inequality satisfaction is best-effort within an iteration budget. Since every correction iterate re-enters the completion model, the returned state is dynamically consistent by construction relative to that model and the supplied previous-state anchor. MaDE drives dynamics residuals to essentially zero on fully specified simulated systems, and on an underspecified system leaves a smaller true-dynamics residual than the baselines. Designed to attach to arbitrary predictors, the frozen operator is evaluated downstream of recurrent, structured state-space, and transformer predictors. On recorded vehicle trajectories the one-step residual against a kinematic bicycle model is 0.0071 to 0.0072 for MaDE and 0.1703 to 0.1714 for raw predictors. MaDE raises average displacement error by a factor of 1.57 to 1.83.
\end{abstract}

\section{Introduction}

\begin{wrapfigure}{r}{0.48\textwidth}
    \centering
    \includegraphics{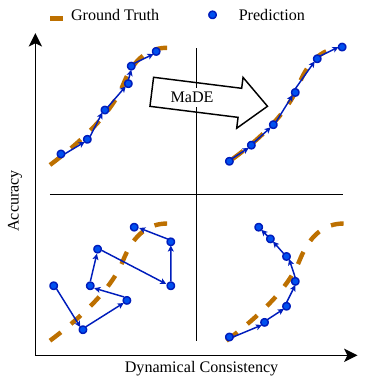}
    \caption{Schematic of MaDE. MaDE improves the dynamical consistency of trajectory predictions at a cost in accuracy.}
    \label{fig:motivation}
\end{wrapfigure}

Neural trajectory models are increasingly used in physical-system pipelines for vehicle trajectory reconstruction, simulation, and behavioural forecasting. In these settings, predicted trajectories do not suffice as merely statistical outputs, since they are consumed by downstream modules that implicitly assume physical admissibility. A trajectory may achieve low prediction error while still violating actuator limits, state constraints, or basic dynamical consistency (Figure~\ref{fig:motivation}), making it unreliable for physical interpretation, safety analysis, or control-aware reasoning.

This mismatch arises because modern neural predictors are typically optimised for predictive accuracy rather than feasibility. Existing approaches partially address the problem by embedding optimisation layers \citep{amos2017optnet,agrawal2019cvxpylayers}, incorporating physics priors \citep{raissi2019pinns,karniadakis2021piml}, or learning structured dynamical models \citep{chen2018neuralode,rackauckas2020ude,yin2021aphynity}. However, these approaches often require retraining the predictor itself, assume analytically specified dynamics, or tightly couple feasibility enforcement to the upstream sequence architecture. In many practical settings, the available physics is incomplete and controls are unobserved.

We introduce \textbf{Ma}rkovian \textbf{D}ynamics \textbf{E}nforcer (MaDE), a differentiable downstream correction operator for neural trajectory predictions. An inverse dynamics model infers latent controls from adjacent states, and a known-physics model augmented with a learned residual uses those controls to recompute the state. A correction stage then adjusts the inferred controls by gradient descent on a violation objective for explicit state and control bounds, while remaining on the learned dynamics manifold. The resulting operator is Markovian, time-invariant, and deterministic, and is designed to attach downstream of arbitrary predictors without retraining them. Dynamics consistency holds by construction relative to the learned completion operator and the supplied previous-state anchor, and asserts nothing about validity against the true data-generating system. Inequality satisfaction by the correction stage is best-effort and limited to a fixed iteration budget.

This work makes the following contributions. We define a post-hoc feasibility-enforcement problem for neural trajectory proposals under partially specified controlled dynamics and unobserved controls. For that problem we instantiate a completion--correction operator whose completion function is a learned controlled dynamics manifold rather than a fully specified analytical equality map. Its correction is posed in physically typed control space from state-only data. We evaluate the operator in two settings, reporting every method's dynamics residuals and inequality violations as measured. On a simulation ladder from fully specified to underspecified dynamics, the operator preserves learned dynamics consistency and improves fidelity over direct-mapping and learned-projection baselines. Downstream of three independently trained neural trajectory predictors on recorded vehicle trajectories, the frozen operator lowers the known-model dynamics residual at an accuracy cost we quantify. Its effect on the inequality violation rate is reported separately from that residual.

\section{Related Work}

\paragraph{Differentiable optimisation and completion--correction.} Optimisation layers embed convex programs in neural networks and differentiate through their solutions \citep{amos2017optnet,agrawal2019cvxpylayers}, while related approaches parameterise feasible sets directly \citep{frerix2020constraints}. \citet{donti2021dc3} introduce a completion--correction framework in which a network proposes free variables, equality constraints determine the remaining variables, and inequality violations are reduced by gradient steps that remain on the equality manifold. Optimisation layers assume an analytical equality map. MaDE is closest to the completion--correction framework, adapting it to an equality manifold that a known physical model and a learned residual supply instead of an analytical specification.

\paragraph{Continuous-time and state-space sequence models.} Neural ODEs and Neural CDEs learn continuous vector fields or controlled differential equations that are integrated forward in time \citep{chen2018neuralode,kidger2020neuralcde}. Kalman filtering provides the classical predict-update structure for sequential estimation \citep{kalman1960filtering}, and deep state-space and latent-sequential models extend the state-space formulation with nonlinear transitions, variational inference, or expressive recurrent parameterisations \citep{krishnan2015dkf,gu2022s4,gu2024mamba}. These methods estimate or generate whole sequences while carrying state across the trajectory. MaDE is a discrete, stateless, per-transition correction operator that projects arbitrary trajectory proposals onto an explicitly constrained learned dynamics manifold.

\paragraph{Imitation from observation.} Here the actions that produced a demonstration are unavailable. Behavioural Cloning from Observation infers missing actions through an inverse dynamics model before policy learning \citep{torabi2018bco}, while related methods match transition distributions adversarially \citep{torabi2018gaifo} or introduce latent action spaces before mapping them to executable controls \citep{edwards2019ilpo}. Inverse-dynamics prediction also supplies self-supervision for representation learning and exploration \citep{pathak2017curiosity}. Imitation from observation recovers missing actions to learn a policy, imitate an expert, or match transition distributions, whereas MaDE's inverse dynamics recovers the physically typed controls the completion step needs.

\paragraph{Hybrid physical models.} Physics-informed neural networks impose physical laws via architectural or training-time constraints \citep{raissi2019pinns,karniadakis2021piml}, while universal differential equations and hybrid simulators couple mechanistic and neural components in differentiable pipelines \citep{rackauckas2020ude,heiden2021neuralsim}. APHYNITY \citep{yin2021aphynity} is the closest comparison among these hybrid models, decomposing dynamics into a known physical component and a minimum-norm learned augmentation. Hybrid physical models fit a learned residual inside an end-to-end autonomous dynamics model, without inferring unobserved controls and without reducing inequality violations. MaDE adds those two steps to the APHYNITY decomposition.

\paragraph{Vehicle trajectory pipelines.} Kinodynamic plausibility in vehicle trajectory pipelines is enforced by classical estimation over a kinematic motion model, by kinematic architectures, and by optimal control. Rauch-Tung-Striebel smoothing adds a fixed-interval backward recursion to the predict-update filtering structure above \citep{rauch1965maximum}. Deep Kinematic Models place a kinematic vehicle model in the predictor's final layer so that decoded trajectories are kinematically feasible \citep{cui2020deepkinematic}. The Realistic Residual Block appends a module carrying kinematic knowledge to a data-driven predictor \citep{bahari2021injecting}. Differentiable model predictive control embeds a finite-horizon optimal control solve as a network layer that is itself the policy, with its cost and dynamics terms learned end to end \citep{amos2018diffmpc}. Each of these four methods assumes an analytically specified motion model, or is a component of the model that produces the trajectory rather than an operator applied to a finished proposal. MaDE infers unobserved physical controls for a supplied transition and reduces analytical inequality violations on the state those controls produce, under dynamics that are only partially specified.

\section{Methodology}
\label{sec:method}

Markovian Dynamics Enforcer (MaDE) is a Markovian, time-invariant one-step correction operator for controlled dynamical systems with unobserved controls. It maps adjacent state estimates from data, simulation, or an upstream predictor to a state-control pair on a learned dynamics manifold, corrected against explicit inequality constraints. Statelessness lets one operator serve any horizon, because the same cell is applied to every transition and no separate model is trained per trajectory length within a system. The cost is that a horizon is corrected by recursive application of a local map, which can trade sequence-level accuracy for one-step dynamic admissibility, at a size Section~\ref{sec:results} reports.

\subsection{Problem Setup}

Let $x_{t-1} \in \mathbb{R}^{n_x}$ denote the supplied previous-state anchor for the one-step correction problem, and let $\tilde{x}_t \in \mathbb{R}^{n_x}$ denote the raw next-state proposal presented to MaDE. The tilde therefore marks the state that is proposed for correction, not the anchor used as the ODE initial condition. Let $u_{t-1} \in \mathbb{R}^{n_u}$ denote the physical control over the interval $[t-1,t]$. The control is not observed during training or inference.
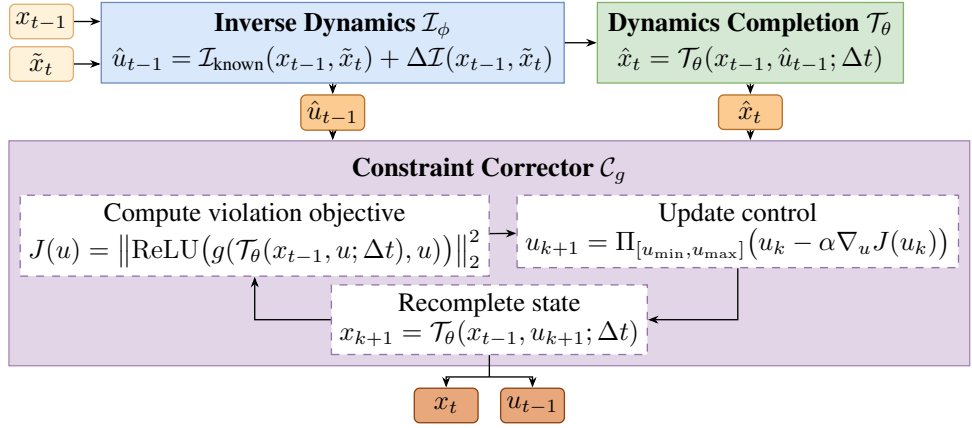
\begin{figure}[t]
    \centering
    \definecolor{madeInFill}{HTML}{FFF2DA}\definecolor{madeInLine}{HTML}{E0B073}
\definecolor{madeMidFill}{HTML}{FBCB91}\definecolor{madeMidLine}{HTML}{B56F2A}
\definecolor{madeOutFill}{HTML}{E8A87C}\definecolor{madeOutLine}{HTML}{8B4513}
\definecolor{madeInvFill}{HTML}{DAE8FC}\definecolor{madeInvLine}{HTML}{6C8EBF}
\definecolor{madeCompFill}{HTML}{D5E8D4}\definecolor{madeCompLine}{HTML}{82B366}
\definecolor{madeCorrFill}{HTML}{E1D5E7}\definecolor{madeCorrLine}{HTML}{9673A6}
\begin{tikzpicture}[
    >={Stealth[length=4pt]}, line width=0.5pt,
    var/.style={draw, rounded corners=2pt, inner sep=2pt, minimum height=13pt, minimum width=24pt},
    madein/.style={var, fill=madeInFill, draw=madeInLine},
    mid/.style={var, fill=madeMidFill, draw=madeMidLine},
    madeout/.style={var, fill=madeOutFill, draw=madeOutLine},
    block/.style={draw, align=center, inner sep=4pt},
    madestep/.style={draw=madeCorrLine, dashed, fill=white, align=center, inner sep=3pt, minimum height=26pt},
  ]
  \node[block, fill=madeInvFill, draw=madeInvLine] (inv)
    {\textbf{Inverse Dynamics $\mathcal{I}_\phi$}\\[2pt]
     $\hat{u}_{t-1} = \mathcal{I}_{\text{known}}(x_{t-1}, \tilde{x}_t) + \Delta\mathcal{I}(x_{t-1}, \tilde{x}_t)$};
  \node[madein, anchor=east] (xprev) at ([xshift=-10pt, yshift=8pt]inv.west) {$x_{t-1}$};
  \node[madein, anchor=east] (xtil) at ([xshift=-10pt, yshift=-8pt]inv.west) {$\tilde{x}_t$};
  \node[block, fill=madeCompFill, draw=madeCompLine, anchor=west] (comp) at ([xshift=12pt]inv.east)
    {\textbf{Dynamics Completion $\mathcal{T}_\theta$}\\[2pt]
     $\hat{x}_t = \mathcal{T}_\theta(x_{t-1}, \hat{u}_{t-1}; \Delta t)$};
  \draw[->] (xprev.east) -- (xprev.east -| inv.west);
  \draw[->] (xtil.east) -- (xtil.east -| inv.west);
  \draw[->] (inv) -- (comp);
  \node[mid, below=4pt of inv] (uhat) {$\hat{u}_{t-1}$};
  \node[mid, below=4pt of comp] (xhat) {$\hat{x}_t$};
  \draw[->] (inv) -- (uhat);
  \draw[->] (comp) -- (xhat);

  \node[madestep, anchor=north west] (obj) at ([xshift=4pt, yshift=-24pt]uhat.south -| xprev.west)
    {Compute violation objective\\
     $J(u) = \bigl\| \mathrm{ReLU}\bigl(g(\mathcal{T}_\theta(x_{t-1}, u; \Delta t), u)\bigr) \bigr\|_2^2$};
  \node[madestep, anchor=north west] (upd) at ([xshift=10pt]obj.north east)
    {Update control\\
     $u_{k+1} = \Pi_{[u_{\min}, u_{\max}]}\bigl(u_k - \alpha \nabla_u J(u_k)\bigr)$};
  \coordinate (corrmid) at ($(obj.north west)!0.5!(upd.north east)$);
  \node[madestep, anchor=north] (rec) at ([yshift=-6pt]corrmid |- upd.south)
    {Recomplete state\\
     $x_{k+1} = \mathcal{T}_\theta(x_{t-1}, u_{k+1}; \Delta t)$};
  \draw[->] (obj) -- (upd);
  \draw[->] (upd.south) |- (rec.east);
  \draw[->] (rec.west) -| (obj.south);
  \node[anchor=south, inner sep=2pt] (hdr) at ([yshift=2pt]$(obj.north west)!0.5!(upd.north east)$)
    {\textbf{Constraint Corrector $\mathcal{C}_g$}};
  \begin{scope}[on background layer]
    \node[draw=madeCorrLine, fill=madeCorrFill, inner sep=4pt, fit=(hdr)(obj)(upd)(rec)] (corr) {};
  \end{scope}
  \draw[->] (uhat) -- (uhat |- corr.north);
  \draw[->] (xhat) -- (xhat |- corr.north);

  \coordinate (split) at ([yshift=-4pt]rec.south |- corr.south);
  \node[madeout, anchor=north east] (xt) at ([xshift=-4pt, yshift=-4pt]split) {$x_t$};
  \node[madeout, anchor=north west] (ut) at ([xshift=4pt, yshift=-4pt]split) {$u_{t-1}$};
  \draw (rec.south) -- (split);
  \draw[->] (split) -| (xt.north);
  \draw[->] (split) -| (ut.north);
\end{tikzpicture}
    \caption{Overview of the MaDE cell. The control update is drawn with zero momentum.}
    \label{fig:made-cell}
\end{figure}

Feasibility is defined by two constraints. First, the returned state must be consistent with a dynamics map $\mathcal{T}_\theta$ over the sampling interval $\Delta t$:
\begin{equation}
    x_t = \mathcal{T}_\theta(x_{t-1}, u_{t-1}; \Delta t).
    \label{eq:made-dynamics-equality}
\end{equation}
Second, the returned transition must satisfy the analytical inequality constraint $g(x_t, u_{t-1}) \leq 0$, where $g$ encodes state bounds, actuator bounds, speed limits, or other one-step local domain constraints. Appendix~\ref{app:e01-systems} instantiates $g$ per system. It reaches the corrector through the violation objective $J$ of Eq.~\eqref{eq:made-correction-objective} and training through the Phase 2 inequality penalty. For a fixed previous state, the target feasible set is therefore the set of controls whose completed next state satisfies both Eq.~\eqref{eq:made-dynamics-equality} and the inequality condition above.

\subsection{The MaDE Cell}

MaDE instantiates a completion-correction pipeline inspired by DC3 \citep{donti2021dc3}, with a learned dynamics model as the completion function (Figure~\ref{fig:made-cell}). A single cell is
\begin{equation}
    \mathcal{M}_{\Theta,g}:
    (x_{t-1}, \tilde{x}_t)
    \mapsto
    (x_t, u_{t-1}),
\end{equation}
with shared parameters $\Theta = \{\phi,\theta\}$. It is composed of three operators:
\begin{align}
    \hat{u}_{t-1}
    &= \mathcal{I}_\phi(x_{t-1}, \tilde{x}_t),
    \label{eq:made-i-step}
    \\
    \hat{x}_t
    &= \mathcal{T}_\theta(x_{t-1}, \hat{u}_{t-1}; \Delta t),
    \label{eq:made-t-step}
    \\
    (x_t, u_{t-1})
    &= \mathcal{C}_g(x_{t-1}, \hat{x}_t, \hat{u}_{t-1}).
    \label{eq:made-c-step}
\end{align}
The inverse dynamics operator $\mathcal{I}_\phi$ proposes the free variables: physically typed controls $\hat{u}_{t-1}$ that explain the transition from the supplied anchor $x_{t-1}$ to the raw proposal $\tilde{x}_t$. The augmented dynamics operator $\mathcal{T}_\theta$ completes the transition by recomputing the next state $\hat{x}_t$ from the anchor $x_{t-1}$ and inferred control $\hat{u}_{t-1}$. The corrector $\mathcal{C}_g$ updates the control to reduce inequality violations, recomputing the state through $\mathcal{T}_\theta$ after every update. Thus the proposal $\tilde{x}_t$ guides control inference, but the output state is generated by the dynamics model rather than copied from the proposal.

The analogy is conceptual, and the mechanism DC3 uses does not carry over to either stage. DC3 completes by solving an implicit algebraic equality system and corrects with gradient steps projected onto the tangent space of the equality manifold \citep{donti2021dc3}. MaDE completes by explicit forward numerical integration, so any state $\mathcal{T}_\theta$ returns satisfies Eq.~\eqref{eq:made-dynamics-equality} without an implicit solve. The corrector re-integrates through $\mathcal{T}_\theta$ after each control update, which does the work a tangent projection does in DC3.

\subsection{Completion and Inverse Dynamics}
\label{sec:method:inverse}

The completion model follows the APHYNITY idea of augmenting a known physical vector field with a minimum-norm learned residual \citep{yin2021aphynity}. For a generic previous state $x_{\mathrm{prev}}$, MaDE uses a continuous-time controlled vector field $\dot{X}(\tau) = f_{\mathrm{phys}}(X(\tau), u_{t-1}) + F_{a,\theta}(X(\tau), u_{t-1})$ with $X(0) = x_{\mathrm{prev}}$, and exposes the fixed-step map $\mathcal{T}_\theta(x_{\mathrm{prev}}, u_{t-1}; \Delta t) = X(\Delta t)$. The known component $f_{\mathrm{phys}}$ supplies the structural dynamics, while $F_{a,\theta}$ captures the part of the dynamics that the known model cannot explain. The residual is trained with a minimum-norm penalty, written schematically as $\mathcal{R}_a = \mathbb{E}[\|F_{a,\theta}(x, u)\|_2^2]$, which discourages it from absorbing structure already represented by the physics. Systems with no useful known physics model are therefore outside the intended scope of the method.

The inverse dynamics operator $\mathcal{I}_\phi$ of Eq.~\eqref{eq:made-i-step} infers a control that explains the transition between adjacent states under the completion model, as a fixed prior plus a learned residual:
\begin{equation}
    \hat{u}_{t-1}
    =
    \mathcal{I}_\phi(x_{t-1}, \tilde{x}_t)
    =
    \mathcal{I}_{\mathrm{known}}(
        x_{t-1}, \tilde{x}_t; \Delta t
    )
    +
    \Delta \mathcal{I}_{\phi}(
        x_{t-1}, \tilde{x}_t
    ).
    \label{eq:made-inverse-decomp}
\end{equation}
The fixed term $\mathcal{I}_{\mathrm{known}}$ is a system-specific kinematic prior, such as a finite-difference acceleration estimate or a closed-form steering approximation. The inferred control is physically typed and bounded, but it is model-relative: it is the control that makes the completion model reproduce the transition, rather than an estimate of the command the system actually applied (Appendix~\ref{app:ctrl-recovery}). The residual $\Delta \mathcal{I}_\phi$ corrects discretisation error and mismatch between the prior and the data. Initialising this residual near zero anchors the model to physically meaningful controls at the start of training, reducing the chance that the inverse model and dynamics residual learn a mutually convenient but physically uninterpretable decomposition. A minimum-norm penalty $\mathcal{R}_{\Delta\mathcal{I}} = \mathbb{E}[\|\Delta \mathcal{I}_\phi(x_{t-1}, \tilde{x}_t)\|_2^2]$, mirroring the one on $F_a$, keeps the inverse residual small throughout training.

\subsection{Control-Space Inequality Correction}

The corrector enforces inequalities by optimising over controls, not by directly clipping the state. For a fixed previous state and parameters, define the violation objective
\begin{equation}
    J(u)
    =
    \left\|
        \mathrm{ReLU}
        \left(
            g\left(
                \mathcal{T}_\theta(x_{t-1}, u; \Delta t),
                u
            \right)
        \right)
    \right\|_2^2.
    \label{eq:made-correction-objective}
\end{equation}
Starting from $\hat{u}_{t-1}$ with $v^0 = 0$, MaDE applies gradient updates with optional momentum:
\begin{equation}
    \begin{aligned}
        v^{k+1}
        &=
        \beta\, v^k
        -
        \alpha \nabla_u J(u^k),
        \\
        u^{k+1}
        &=
        \Pi_{[u_{\min}, u_{\max}]}\left(u^k + v^{k+1}\right),
        \\
        x^{k+1}
        &= \mathcal{T}_\theta(x_{t-1}, u^{k+1}; \Delta t),
    \end{aligned}
    \label{eq:made-correction-update}
\end{equation}
where $\beta \in [0,1)$ is the momentum coefficient ($\beta = 0$ recovers
projected gradient descent) and
$\Pi_{[u_{\min}, u_{\max}]}(z)=\min(\max(z,u_{\min}),u_{\max})$ is the
componentwise projection onto the control box defined by the inequality
constraints.
The re-completion step in Eq.~\eqref{eq:made-correction-update} is essential, since every intermediate correction iterate remains on the dynamics manifold of $\mathcal{T}_\theta$ and the algorithm never edits $x_t$ independently of the dynamics. The correction loop runs a fixed number of steps at training time and an adaptive, tolerance-based number at evaluation time, detailed in Appendix~\ref{app:solvers:optimisation}.

\subsection{Training Objectives}

MaDE is trained by self-supervision on state transitions alone, with no ground-truth controls: the inverse and forward operators supervise each other through cycle consistency, and the augmented dynamics residual is kept small through the minimum-norm penalties. Training runs in two phases, summarised in Algorithm~\ref{alg:made-training} in Appendix~\ref{app:solvers}: Phase 1 fits the consistency losses and residual penalties, and Phase 2 adds inequality-aware training through the corrector. In both phases the observed next state $x_t$ is used as the proposal input in place of $\tilde{x}_t$, so the loss definitions below reuse feasible dataset transitions while keeping the same one-step operator structure.
For a data transition $(x_{t-1},x_t)$, forward consistency trains the inferred control and completed dynamics to reconstruct the observed next state:
\begin{equation}
    \mathcal{L}_{\mathrm{fwd}}
    =
    \left\|
        \mathcal{T}_\theta
        \left(
            x_{t-1},
            \mathcal{I}_\phi(x_{t-1}, x_t); \Delta t
        \right)
        -
        x_t
    \right\|_2^2.
    \label{eq:made-forward-loss}
\end{equation}
Inverse consistency samples a control $u \sim q(u)$, completes a synthetic transition through the current dynamics, and trains the inverse model to recover the sampled control:
\begin{equation}
    \mathcal{L}_{\mathrm{inv}}
    =
    \left\|
        \mathcal{I}_\phi
        \left(
            x_{t-1},
            \mathcal{T}_\theta(x_{t-1}, u; \Delta t)
        \right)
        -
        u
    \right\|_2^2.
    \label{eq:made-inverse-loss}
\end{equation}
The two terms supervise complementary failure modes. $\mathcal{L}_{\mathrm{fwd}}$ tests whether the inferred control reproduces the next state through $\mathcal{T}_\theta$, and used alone leaves the inverse map free to drift toward a mutually convenient but physically opaque decomposition with $\mathcal{T}_\theta$. $\mathcal{L}_{\mathrm{inv}}$ tests the reverse, recovering controls from synthetic transitions produced by $\mathcal{T}_\theta$ without requiring $\mathcal{I}_\phi$ to explain the observed data.

The total Phase 1 loss $\mathcal{L}_1 = \mathcal{L}_{\mathrm{fwd}} + \lambda_{\mathrm{inv}}\mathcal{L}_{\mathrm{inv}} + \lambda_a \mathcal{R}_a + \lambda_{\Delta\mathcal{I}} \mathcal{R}_{\Delta\mathcal{I}}$ is logged for validation, early stopping and comparison across alternation schedules and stopping criteria, and no module is trained directly against it. Algorithm~\ref{alg:made-training} in Appendix~\ref{app:solvers} gives the per-side objectives and the gradient-freezing rule that keeps the inverse model and augmented dynamics from freely colluding while still letting the two modules improve each other over training.

Phase 2 introduces inequality-aware training through the full pipeline, applying the corrector to the training pair to give $(x_t^{\mathcal{C}}, u_{t-1}^{\mathcal{C}}) = \mathcal{C}_g\left(x_{t-1}, \mathcal{T}_\theta\left(x_{t-1}, \mathcal{I}_\phi(x_{t-1}, x_t); \Delta t\right), \mathcal{I}_\phi(x_{t-1}, x_t)\right)$. The inequality loss is
\begin{equation}
    \mathcal{L}_{\mathrm{ineq}}
    =
    \left\|
        \mathrm{ReLU}
        \left(
            g(x_t^{\mathcal{C}}, u_{t-1}^{\mathcal{C}})
        \right)
    \right\|_2^2,
    \label{eq:made-ineq-loss}
\end{equation}
and the total Phase 2 loss $\mathcal{L}_2 = \mathcal{L}_1 + \lambda_g \mathcal{L}_{\mathrm{ineq}}$ is likewise only logged, with Phase 2 updates alternating under the same freezing rule. The inequality term updates the inverse model: it shapes $\mathcal{I}_\phi$ toward controls that already lie closer to the feasible region, reducing the amount of correction needed at inference time. Its gradient is stopped in the loss that updates $\mathcal{T}_\theta$, so the forward model and its learned residual $F_a$ receive no inequality signal. The corrector itself remains an algorithmic layer rather than a separate learned network.

\section{Experimentation}
\label{sec:experiments}

MaDE is evaluated in two experiments: a simulated four-system ladder spanning fully specified and underspecified known physics in Section~\ref{sec:experiments:e01}, and the real-data experiment on inD in Section~\ref{sec:experiments:e05}. The two share the evaluation metrics and the training protocol of Section~\ref{sec:experiments:metrics}.

\subsection{Evaluation metrics and training protocol}
\label{sec:experiments:metrics}

We report both the rate and the magnitude of inequality violations. The rate (Ineq.\ rate) is the fraction of steps violating the analytical bounds $g \leq 0$, and the magnitude (Ineq.\ mag.)\ is the mean violation norm over the violating steps. Both score states and controls together, on a method's own controls where it emits them and otherwise on controls recovered from its states through the inverse of the known model. The one-step residual of the returned transition is measured against the known physics (Dyn.-K), the augmented dynamics (Dyn.-L), and the true generating model (Dyn.-T). Fidelity (Fid.)\ is the mean Euclidean distance from the corrected next state to the unperturbed simulator trajectory at the same step. The average displacement error (ADE) is the mean Euclidean distance in metres from the returned position $(x, y)$ to the recorded position over the prediction horizon. The final displacement error (FDE) is that distance at the last step. Fid.\ and ADE both measure distance to a reference, and the two experiments supply different ones. Fid.\ scores each corrected transition against the unperturbed simulator state, and ADE scores the whole forecast against the recorded trajectory, the only reference recorded data provide. Table~\ref{tab:metrics} in Appendix~\ref{app:e01-systems} gives every metric except ADE and FDE in symbols, and Appendix~\ref{app:solvers:optimisation} gives MaDE's training protocol, validation-driven early stopping in each phase.

\subsection{Single-agent simulated dynamics}
\label{sec:experiments:e01}

We work with a single agent, fixed scalar physics parameters, and no upstream predictor, so the pipeline reduces to the bare MaDE cell of Section~\ref{sec:method}. Access to the true generating process makes this the controlled setting of the paper. Posing correction on a single state transition lets MaDE sit behind a predictor with no access to its history, horizon, or decoding scheme, and items 1 and 2 of Appendix~\ref{app:scope-failure} state what this scope puts out of reach.

\paragraph{Data and seeds.} Training data consist of feasible trajectories of length $T = 32$ generated by integrating the true system over the sampling interval $\Delta t = 0.1$. Each cell is run for five random seeds, each a complete training run of every learned method. Appendix~\ref{app:e01-systems} gives the splits and what each seed varies. At evaluation time every proposal after the unperturbed feasible state $x_0$ is perturbed under the condition-specific protocol below, and MaDE is required to project the perturbed transitions back into the feasible set. Within a seed every method and ablation sees the same perturbed proposals, so the comparisons are paired. Correction is applied recursively along the trajectory from $x_0$, and each corrected state is the anchor for the next transition over the whole horizon.

\paragraph{Systems.} We evaluate MaDE under four conditions, linear double integrator (\textbf{DI}), kinematic unicycle (\textbf{UNI}), kinematic bicycle (\textbf{KB}), and dynamic bicycle (\textbf{DB}), each specified in Appendix~\ref{app:e01-systems}. For each of DI, UNI, and KB, the known forward dynamics $\mathcal{T}_{\mathrm{phys}}$ and analytical inverse prior $\mathcal{I}_{\mathrm{known}}$ match the data-generating model. Dyn.-T therefore equals Dyn.-K on these three systems, and the dynamics decomposition predicts $F_a \to 0$ at convergence, which reduces the augmented dynamics of Dyn.-L to the known physics. These three systems act as pipeline-correctness checks at progressively richer geometry, and as a control that the residual does not absorb structure already explained by the known physics. Table~\ref{tab:e01-results} reports Dyn.-K alone for them, and Table~\ref{tab:e01-ablations} in Appendix~\ref{app:e01-ablations} reports all three for MaDE and its variants. For DB, the data-generating model is the dynamic bicycle, while MaDE retains the kinematic bicycle as both $\mathcal{T}_{\mathrm{phys}}$ and $\mathcal{I}_{\mathrm{known}}$. The forward residual $F_a$ and inverse residual $\Delta\mathcal{I}_\phi$ must therefore absorb the missing tyre forces, lateral slip, and inertial coupling. All four conditions take a deterministic bound-violation evaluation perturbation of scale $0.10$, with angular state components at half scale, $0.05$, where present. At evaluation, DB inputs also carry zero-mean Gaussian observation noise of scale $0.02$. For training on DB, where the known dynamics are underspecified, MaDE adds a separate zero-mean Gaussian augmentation of scale $0.02$ to both states of each training and validation transition. The fully specified systems train without augmentation.

\paragraph{Baselines.} The baseline set is clamp-only, per-step MLP, and FAB, which span the natural alternatives to MaDE's completion-correction structure. \textbf{Clamp-only} enforces available box constraints by elementwise clipping and ignores dynamics consistency, isolating what a constraint-aware but physics-unaware projection achieves. \textbf{Per-step MLP} directly maps a proposed transition to a corrected transition with no explicit physics or controls, isolating what an unstructured one-step regressor learns from feasible data. \textbf{FAB} follows Fast Autoencoder-Based Projections \citep{chzhen2026fab}: a transition-pair encoder, a clip to a fixed-radius latent ball, and a single decoder return an approximate feasible projection, giving a structured learned-projection baseline without explicit physics, controls, or dynamics-manifold completion. We reimplement both of its training phases (Appendix~\ref{app:solvers:baselines}).

\paragraph{Scoring.} The simulated methods fall into two groups for scoring. MaDE, its variants and the prior-only comparator of Appendix~\ref{app:prior-only} emit a control with each state, and the three baselines emit none. Dyn.-K scores every method on controls recovered from its returned states through the inverse of the known model. On DB, Dyn.-T scores MaDE, its variants and the prior-only comparator on their emitted controls, and scores each baseline on controls recovered through the inverse of the known kinematic bicycle. Dyn.-L scores MaDE and its variants on their emitted controls under their own learned dynamics. The baselines have no learned dynamics or inverse model, so a baseline's Dyn.-L is taken under the MaDE model trained on the same system and seed (Appendix~\ref{app:e01-ablations}). A baseline's Dyn.-L therefore measures agreement with MaDE's model rather than a model of its own. The inequality metrics score the methods that emit controls on those controls, and each baseline on recovered controls.

\subsection{Correction of neural trajectory predictors}
\label{sec:experiments:e05}

We train three physics-unaware predictors from scratch, a recurrent model (LSTM) \citep{hochreiter1997lstm}, a compact structured state-space (S6/Mamba-style) sequence model \citep{gu2024mamba}, and a transformer \citep{vaswani2017attention}, each at five seeds, giving 15 predictor runs. Trajectories are sampled at 5 Hz, a step of $\Delta t = 0.2$ s, and each predictor reads a 10-step (2 s) state history and per-vehicle metadata. It decodes the 15-step (3 s) horizon in one pass, as cumulative state deltas from the last observed state. All three are trained on a mean-squared-error objective alone, with no dynamics penalty and no inequality penalty. MaDE therefore corrects the predictor's own forecast error rather than a synthetic perturbation.

\paragraph{Data and MaDE training.} In the real-data experiment the windows come from the inD dataset of drone-recorded intersections \citep{bock2020ind}. MaDE is trained on inD states and per-vehicle metadata, with no control labels, under the training protocol of Section~\ref{sec:experiments:metrics}, then frozen. Its known model is the kinematic bicycle at a reference wheelbase of 2.7 m. Its inverse and completion steps add to that wheelbase a signed per-vehicle residual, which a small encoder trained with $F_a$ predicts from the vehicle's recorded length, width, class and recording location. Dyn.-K scores every method on controls recovered through the inverse of the known model at the 2.7 m reference wheelbase, without the per-vehicle wheelbase residual that MaDE learns. The recordings are dominated by vehicles that never move, and two displacement filters remove them: a window-level filter for the predictors and the evaluation, and a track-level filter for MaDE, which trains on transition pairs and builds no prediction windows. Both use a displacement threshold of 0.5 m, and Appendix~\ref{app:e01-systems} defines them and gives the counts each leaves. MaDE is trained at three seeds, and each of the three MaDE models corrects all 15 predictor runs, giving 45 cells that separate predictor variation from MaDE's own.

\paragraph{Constraints.} Every method of the real-data experiment is scored against one physical constraint set, the set MaDE is trained against, whose bounds were set by hand as reasonable limits (Appendix~\ref{app:e01-systems}). The clamp baseline projects onto the state bounds of that same set.

Table~\ref{tab:e05-ind} reports the combined inequality violation rate, in which a step counts as violating when it breaks any state or control bound. Appendix~\ref{app:ineq-breakdown} splits the rate and the magnitude into their state-bound and control-bound components for every method, and states how the components combine. The split matters for clamp, which satisfies the state bounds by construction but cannot reach the control bounds, so all of its violation lies on the controls.

\paragraph{Baselines.} The real-data experiment sets MaDE against three baselines, each built on the same predictor forecast that MaDE corrects. The raw forecast itself serves as the uncorrected reference, and the clamp-only baseline of Section~\ref{sec:experiments:e01} projects it onto the state bounds. An EKF/RTS smoother, the classical engineering answer to post-hoc kinematic feasibility, completes the set, and it takes each forecast whole where MaDE corrects one transition pair at a time. It runs an extended Kalman filter (EKF) forward pass \citep{kalman1960filtering,jazwinski1970filtering} followed by a fixed-interval Rauch-Tung-Striebel (RTS) backward pass \citep{rauch1965maximum}. Its process model is MaDE's known model, the 2.7 m kinematic bicycle without the per-vehicle wheelbase residual, so the smoother carries that model's misspecification on real vehicles. None of the inequality constraints MaDE is trained against enters the smoother, whose controls are those it infers within its estimated state. For the inequality metrics, MaDE and the smoother are scored on their own controls, and the raw predictor and clamp on controls recovered from their states. The smoother's covariances are tuned on the training split (Appendix~\ref{app:solvers:baselines}), where the lowest ADE picks the accuracy-tuned setting and the lowest residual against the known model picks the residual-tuned one.

\section{Results and Discussion}
\label{sec:results}

Section~\ref{sec:results:sim} scores the bare operator on the simulated ladder, where the true generating model is known and no upstream predictor intervenes. Section~\ref{sec:results:ind} places MaDE behind trained predictors on recorded inD trajectories, which supply no true model. It asks what correction buys there in dynamics residual and inequality violation, and what it costs in accuracy.

\subsection{Simulated experiment}
\label{sec:results:sim}

Table~\ref{tab:e01-results} reports the mean and the population standard deviation over five seeds, with the ablation study, the prior-only comparator and control recovery in Appendix~\ref{app:sim-results}.

\begin{table}[t]
\caption{Results on the four simulated systems. Lower is better for every metric. Within each metric row the best mean is set in \textbf{bold} and the second-best in \underline{underline}.}
\label{tab:e01-results}
\centering
\setlength{\tabcolsep}{3pt}
\renewcommand{\arraystretch}{0.85}
\begin{adjustbox}{max width=\linewidth}\begin{tabular}{llcccc}
\toprule
 & & \multicolumn{3}{c}{Baselines} & \multicolumn{1}{c}{MaDE} \\
\cmidrule(lr){3-5}\cmidrule(lr){6-6}
Cond. & Metric & Clamp & MLP & FAB & MaDE \\
\midrule
        DI & Ineq. rate & \underline{$0.2551\,{\scriptscriptstyle\pm}\,0.0000$} & $0.3955\,{\scriptscriptstyle\pm}\,0.0045$ & $0.3000\,{\scriptscriptstyle\pm}\,0.0129$ & $\mathbf{0.2034\,{\scriptscriptstyle\pm}\,0.0000}$ \\
         & Ineq. mag. & $9.8777\,{\scriptscriptstyle\pm}\,0.0000$ & $4.3835\,{\scriptscriptstyle\pm}\,0.2038$ & \underline{$2.4171\,{\scriptscriptstyle\pm}\,0.4272$} & $\mathbf{0.2428\,{\scriptscriptstyle\pm}\,0.0000}$ \\
         & Dyn.-K & $0.3332\,{\scriptscriptstyle\pm}\,0.0000$ & \underline{$0.2397\,{\scriptscriptstyle\pm}\,0.0012$} & $0.3304\,{\scriptscriptstyle\pm}\,0.0206$ & $\mathbf{0.0000\,{\scriptscriptstyle\pm}\,0.0000}$ \\
         & Fid. & $3.0207\,{\scriptscriptstyle\pm}\,0.0000$ & $2.9800\,{\scriptscriptstyle\pm}\,0.0054$ & \underline{$2.7139\,{\scriptscriptstyle\pm}\,0.1339$} & $\mathbf{2.2792\,{\scriptscriptstyle\pm}\,0.0000}$ \\
        \\[-0.4ex]
        UNI & Ineq. rate & $\mathbf{0.0640\,{\scriptscriptstyle\pm}\,0.0000}$ & $0.3831\,{\scriptscriptstyle\pm}\,0.0038$ & $0.2933\,{\scriptscriptstyle\pm}\,0.0090$ & \underline{$0.1042\,{\scriptscriptstyle\pm}\,0.0000$} \\
         & Ineq. mag. & $4.7181\,{\scriptscriptstyle\pm}\,0.0000$ & $2.1585\,{\scriptscriptstyle\pm}\,0.0553$ & \underline{$1.1305\,{\scriptscriptstyle\pm}\,0.1514$} & $\mathbf{0.0538\,{\scriptscriptstyle\pm}\,0.0000}$ \\
         & Dyn.-K & $0.3251\,{\scriptscriptstyle\pm}\,0.0000$ & \underline{$0.2614\,{\scriptscriptstyle\pm}\,0.0057$} & $0.2729\,{\scriptscriptstyle\pm}\,0.0386$ & $\mathbf{0.0000\,{\scriptscriptstyle\pm}\,0.0000}$ \\
         & Fid. & $5.3315\,{\scriptscriptstyle\pm}\,0.0000$ & $5.4588\,{\scriptscriptstyle\pm}\,0.0128$ & \underline{$5.0115\,{\scriptscriptstyle\pm}\,0.1034$} & $\mathbf{1.4126\,{\scriptscriptstyle\pm}\,0.0000}$ \\
        \\[-0.4ex]
        KB & Ineq. rate & \underline{$0.3008\,{\scriptscriptstyle\pm}\,0.0000$} & $0.5052\,{\scriptscriptstyle\pm}\,0.0078$ & $0.4293\,{\scriptscriptstyle\pm}\,0.0213$ & $\mathbf{0.1965\,{\scriptscriptstyle\pm}\,0.0000}$ \\
         & Ineq. mag. & $5.3723\,{\scriptscriptstyle\pm}\,0.0000$ & $2.8344\,{\scriptscriptstyle\pm}\,0.2041$ & \underline{$1.9033\,{\scriptscriptstyle\pm}\,0.3820$} & $\mathbf{0.1431\,{\scriptscriptstyle\pm}\,0.0001}$ \\
         & Dyn.-K & $0.8546\,{\scriptscriptstyle\pm}\,0.0000$ & $0.7418\,{\scriptscriptstyle\pm}\,0.0155$ & \underline{$0.7152\,{\scriptscriptstyle\pm}\,0.1115$} & $\mathbf{0.0003\,{\scriptscriptstyle\pm}\,0.0000}$ \\
         & Fid. & $13.3756\,{\scriptscriptstyle\pm}\,0.0000$ & $13.7249\,{\scriptscriptstyle\pm}\,0.0424$ & \underline{$13.1332\,{\scriptscriptstyle\pm}\,0.3599$} & $\mathbf{3.6797\,{\scriptscriptstyle\pm}\,0.0006}$ \\
        \\[-0.4ex]
        DB & Ineq. rate & \underline{$0.1595\,{\scriptscriptstyle\pm}\,0.0043$} & $0.4371\,{\scriptscriptstyle\pm}\,0.0189$ & $0.7480\,{\scriptscriptstyle\pm}\,0.0697$ & $\mathbf{0.1384\,{\scriptscriptstyle\pm}\,0.0055}$ \\
         & Ineq. mag. & $12.6677\,{\scriptscriptstyle\pm}\,0.1207$ & \underline{$2.2624\,{\scriptscriptstyle\pm}\,0.1080$} & $6.6766\,{\scriptscriptstyle\pm}\,2.8241$ & $\mathbf{0.2607\,{\scriptscriptstyle\pm}\,0.0392}$ \\
         & Dyn.-K & $1.6078\,{\scriptscriptstyle\pm}\,0.0003$ & \underline{$1.4074\,{\scriptscriptstyle\pm}\,0.0577$} & $2.1756\,{\scriptscriptstyle\pm}\,0.4463$ & $\mathbf{0.0733\,{\scriptscriptstyle\pm}\,0.0326}$ \\
         & Dyn.-L & $1.5567\,{\scriptscriptstyle\pm}\,0.0351$ & \underline{$1.3675\,{\scriptscriptstyle\pm}\,0.0723$} & $2.1809\,{\scriptscriptstyle\pm}\,0.4518$ & $\mathbf{0.0000\,{\scriptscriptstyle\pm}\,0.0000}$ \\
         & Dyn.-T & $1.6976\,{\scriptscriptstyle\pm}\,0.0058$ & \underline{$1.4897\,{\scriptscriptstyle\pm}\,0.0576$} & $2.3552\,{\scriptscriptstyle\pm}\,0.4058$ & $\mathbf{0.3736\,{\scriptscriptstyle\pm}\,0.0647}$ \\
         & Fid. & $13.5112\,{\scriptscriptstyle\pm}\,0.0003$ & $13.8213\,{\scriptscriptstyle\pm}\,0.0264$ & \underline{$12.5074\,{\scriptscriptstyle\pm}\,3.2837$} & $\mathbf{5.2021\,{\scriptscriptstyle\pm}\,0.2740}$ \\
\bottomrule
\end{tabular}
\end{adjustbox}%
\end{table}

\paragraph{Dynamics consistency.} MaDE's returned state is $\mathcal{T}_\theta$ evaluated at the returned control whether or not the corrector iterates (Appendix~\ref{app:scope-failure}), so its Dyn.-L on DB is below $10^{-15}$. On DI, UNI and KB the decomposition predicts that the learned model reduces to the known physics. There MaDE is the only method that drives Dyn.-K to essentially zero, below $0.0004$ on each. On the underspecified DB it has the lowest Dyn.-K and the lowest Dyn.-T of the four methods. The baselines' Dyn.-L there is below their Dyn.-K for clamp-only and the per-step MLP and above it for FAB.

\paragraph{Inequality violation.} Inequality satisfaction rests on the corrector. Without it MaDE has a higher Ineq.\ rate and a higher Ineq.\ mag.\ on all four systems (Appendix~\ref{app:e01-ablations}). With it MaDE has the lowest Ineq.\ mag.\ on all four and the lowest Ineq.\ rate on DI, KB and DB. The violations that remain indicate that the corrector does not always eliminate inequality error within its iteration budget. Clamp-only has the lowest Ineq.\ rate on UNI, yet the highest Ineq.\ mag.\ of the four methods on every system.

\paragraph{Fidelity.} MaDE also has the lowest fidelity error on all four systems, and FAB is second-best on all four, ahead of both clamp-only and the per-step MLP. On DB, the one underspecified system, MaDE's fidelity error is $5.20$ against $12.51$ for FAB.

\paragraph{The FAB baseline.} Away from DB, FAB is second-best on Ineq.\ mag., and on Dyn.-K it is second-best on KB and behind the per-step MLP on DI and UNI. On DB, by contrast, it has the highest Dyn.-K, Dyn.-T and Ineq.\ rate of the four methods. Its population standard deviation across seeds is also larger there than on the other three systems on Ineq.\ rate, Ineq.\ mag., Dyn.-K and fidelity.

\subsection{Real-data experiment}
\label{sec:results:ind}

Table~\ref{tab:e05-ind} compares the four methods within each of the three predictor families, recurrent, state-space and transformer, on the same forecasts, with each cell scored by its mean over the evaluation windows. The table's smoother is the residual-tuned one, because the accuracy-tuned smoother's entries differ only in the state-space family, where its means are higher on all four metrics (Appendix~\ref{app:solvers:baselines}).

\begin{table}[t]
\caption{Results on the inD experiment, as the mean $\pm$ the population standard deviation over cells, 5 per predictor family for each baseline and 15 for MaDE. Lower is better. Within each family and metric, bold marks the lowest mean and underline the second lowest, and ties share a mark.}
\label{tab:e05-ind}
\centering
\renewcommand{\arraystretch}{0.88}
\begin{adjustbox}{max width=\linewidth}\begin{adjustbox}{max width=458.74pt}
{\setlength{\tabcolsep}{2pt}
\begin{tabular}{llcccc}
\toprule
Predictor & Metric & raw & clamp & smoother (res.) & MaDE  \\
\midrule
Recurrent & ADE (m) & $\mathbf{0.6345\,{\scriptscriptstyle\pm}\,0.0240}$ & $\mathbf{0.6345\,{\scriptscriptstyle\pm}\,0.0240}$ & \underline{$0.7122\,{\scriptscriptstyle\pm}\,0.0192$} & $1.1599\,{\scriptscriptstyle\pm}\,0.1367$  \\
 & FDE (m) & $\mathbf{1.7206\,{\scriptscriptstyle\pm}\,0.0586}$ & $\mathbf{1.7206\,{\scriptscriptstyle\pm}\,0.0586}$ & \underline{$1.8096\,{\scriptscriptstyle\pm}\,0.0592$} & $3.0459\,{\scriptscriptstyle\pm}\,0.3766$  \\
 & Dyn.-K & $0.1711\,{\scriptscriptstyle\pm}\,0.0224$ & $0.1687\,{\scriptscriptstyle\pm}\,0.0218$ & \underline{$0.0277\,{\scriptscriptstyle\pm}\,0.0012$} & $\mathbf{0.0072\,{\scriptscriptstyle\pm}\,0.0035}$  \\
 & Ineq. rate & $0.0572\,{\scriptscriptstyle\pm}\,0.0093$ & \underline{$0.0468\,{\scriptscriptstyle\pm}\,0.0091$} & $0.0855\,{\scriptscriptstyle\pm}\,0.0093$ & $\mathbf{0.0114\,{\scriptscriptstyle\pm}\,0.0059}$  \\
\\[-0.4ex]
State-space & ADE (m) & $\mathbf{0.6251\,{\scriptscriptstyle\pm}\,0.0134}$ & $\mathbf{0.6251\,{\scriptscriptstyle\pm}\,0.0134}$ & \underline{$0.7022\,{\scriptscriptstyle\pm}\,0.0178$} & $1.1417\,{\scriptscriptstyle\pm}\,0.0839$  \\
 & FDE (m) & $\mathbf{1.6546\,{\scriptscriptstyle\pm}\,0.0259}$ & $\mathbf{1.6546\,{\scriptscriptstyle\pm}\,0.0259}$ & \underline{$1.7717\,{\scriptscriptstyle\pm}\,0.0290$} & $3.0569\,{\scriptscriptstyle\pm}\,0.2233$  \\
 & Dyn.-K & $0.1714\,{\scriptscriptstyle\pm}\,0.0150$ & $0.1685\,{\scriptscriptstyle\pm}\,0.0145$ & \underline{$0.0277\,{\scriptscriptstyle\pm}\,0.0007$} & $\mathbf{0.0072\,{\scriptscriptstyle\pm}\,0.0035}$  \\
 & Ineq. rate & $0.0583\,{\scriptscriptstyle\pm}\,0.0057$ & \underline{$0.0478\,{\scriptscriptstyle\pm}\,0.0066$} & $0.0849\,{\scriptscriptstyle\pm}\,0.0054$ & $\mathbf{0.0111\,{\scriptscriptstyle\pm}\,0.0053}$  \\
\\[-0.4ex]
Transformer & ADE (m) & $\mathbf{0.7033\,{\scriptscriptstyle\pm}\,0.0175}$ & $\mathbf{0.7033\,{\scriptscriptstyle\pm}\,0.0175}$ & \underline{$0.7776\,{\scriptscriptstyle\pm}\,0.0210$} & $1.1058\,{\scriptscriptstyle\pm}\,0.1466$  \\
 & FDE (m) & $\mathbf{1.7802\,{\scriptscriptstyle\pm}\,0.0304}$ & $\mathbf{1.7802\,{\scriptscriptstyle\pm}\,0.0304}$ & \underline{$1.8652\,{\scriptscriptstyle\pm}\,0.0287$} & $2.9465\,{\scriptscriptstyle\pm}\,0.3834$  \\
 & Dyn.-K & $0.1703\,{\scriptscriptstyle\pm}\,0.0233$ & $0.1671\,{\scriptscriptstyle\pm}\,0.0228$ & \underline{$0.0285\,{\scriptscriptstyle\pm}\,0.0027$} & $\mathbf{0.0071\,{\scriptscriptstyle\pm}\,0.0035}$  \\
 & Ineq. rate & $0.0358\,{\scriptscriptstyle\pm}\,0.0096$ & \underline{$0.0288\,{\scriptscriptstyle\pm}\,0.0080$} & $0.0648\,{\scriptscriptstyle\pm}\,0.0134$ & $\mathbf{0.0065\,{\scriptscriptstyle\pm}\,0.0047}$  \\
\bottomrule
\end{tabular}
}
\end{adjustbox}
\end{adjustbox}%
\end{table}

\paragraph{Dynamics residual.} On inD, MaDE's dynamics consistency holds relative to its learned operator, at the learned per-vehicle wheelbase, and to the supplied anchor. Dyn.-K scores it instead against the kinematic bicycle at the 2.7 m reference wheelbase. On that measure MaDE is lowest in every predictor family, from 0.00712 to 0.00722, against 0.1703 to 0.1714 for the raw predictors (Appendix~\ref{app:per-model}). The residual-tuned smoother is next lowest on Dyn.-K in every family. The recordings themselves, taken from real vehicles at 5 Hz, set the scale for this residual. They carry sensor and tracking noise and follow dynamics the kinematic bicycle only approximates, so none of them sits on the manifold of the 2.7 m kinematic bicycle. With controls recovered from their states in the same way, the recordings score 0.04038 on the 6,681 evaluation windows. On the same windows the raw forecasts score above that level in every family and MaDE scores below it (Table~\ref{tab:e05-ind}). The corrected trajectories therefore sit closer to the 2.7 m kinematic bicycle than the recordings do.

\paragraph{Accuracy cost.} MaDE's correction also costs position accuracy, leaving it with the highest mean FDE of the four methods in every family. Across the three families its ADE rises by a factor of 1.57 to 1.83 over the 3 s horizon, and its FDE by a factor of 1.66 to 1.85 (Appendix~\ref{app:per-model}). Clamp has the same ADE and FDE as the raw predictor in every family, because the position channels carry no bound and clamp leaves every forecast position unchanged. Real data supplies neither true controls nor a true model, so the result is a trade-off between consistency and accuracy under an approximate prior. It is no evidence that the corrected trajectories are physically truer than the recordings.

\paragraph{Inequality violation.} On recorded data, the gap between MaDE's inequality violation rate and the raw predictor's comes from the control bounds. The raw predictors violate the control bounds more often than the state bounds in every family. Every control-bound violation of the raw predictor and of clamp is at the steering-angle bound. MaDE's emitted controls violate a control bound in 2 of its 45 cells, at a rate below $10^{-5}$ in each, so its state-bound rates match its combined rates to four decimal places (Appendix~\ref{app:ineq-breakdown}). Its state-bound rate is above the raw predictor's in the recurrent and state-space families and below it in the transformer family. In every family the combined rate orders MaDE, clamp and the raw predictor from lowest to highest. Both smoothers exceed the raw predictor on the combined rate, on the state bounds and, through their inferred controls, on the control bounds. On recovered controls the recordings have a rate of 0.0047 on the same windows, the rate this scoring assigns to real driving.

\paragraph{Completion and correction.} Completion alone raises inequality violation, and the corrector brings it below the raw forecast. The completion-only variant applies the inverse and completion steps of Eqs.~\eqref{eq:made-i-step} and \eqref{eq:made-t-step} and skips the corrector of Eq.~\eqref{eq:made-c-step}. Each of its 45 cells is paired with the raw forecast of its own predictor family and seed (Appendix~\ref{app:completion-only}). On the evaluation windows the variant raises the inequality violation rate from 0.0504 for the raw forecast to 0.1178, against 0.0096 for the full operator. In all 45 cells the completion-only variant exceeds its paired raw forecast in both violation rate and magnitude, on the evaluation windows and on the unfiltered set of all test windows. The full operator is below the raw forecast on both metrics in all 45 cells on both sets.

\paragraph{Runtime.} Amortised in a batch of 16, MaDE costs a mean of 2.5127 ms per corrected trajectory over the 45 cells of the real-data grid, rising to 4.9297 ms at batch size 1. In both forms the mean exceeds the median because the cells split into a larger cheaper group and a smaller costlier one (Appendix~\ref{app:runtime}). Across cells, the spread at batch size 1 reflects which cells hit the corrector's 50-iteration cap on the timed window. Every cell is timed on that same window, so the spread cannot come from variation in the data. These timings are taken at the $0.2$ s step size of the inD evaluation windows. They come from one run on a device that was idle when the run started, while a second device on the same machine was carrying another process.

The corrector's iteration counts measure the finite-budget caveat of Appendix~\ref{app:scope-failure} when pooled over all 6,681 evaluation windows and the 45 cells, a different population from the single timed window. One call corrects one timestep of one window, and the iteration count per call has a 95th percentile of 1 (Appendix~\ref{app:runtime}). The corrector reaches the 50-iteration cap with the maximum violation still above the $10^{-6}$ tolerance on 0.964\% of calls.

\section{Conclusion}

We introduced MaDE, a time-invariant Markov correction operator that composes inverse dynamics, a known-physics-plus-residual completion model, and a control-space inequality corrector into a differentiable one-step map. Trained from feasible transitions without ground-truth controls, it is a frozen plug-and-play layer, attachable downstream of an arbitrary upstream predictor. Its dynamics consistency holds by construction, relative to the learned completion operator and the supplied previous-state anchor. Its inequality satisfaction, by contrast, is best-effort and limited by the corrector's iteration budget. Across a four-system fully-specified-to-underspecified APHYNITY ladder with deterministic bound-violation and Gaussian observation-noise stress tests, MaDE achieves better trajectory fidelity than learned-projection and direct-mapping baselines. In the inD experiment it gives the lowest one-step residual against the 2.7 m kinematic bicycle and the lowest inequality violation rate of the compared methods in every predictor family. The accuracy cost measured there is a rise in average displacement error by a factor of 1.57 to 1.83 over the raw predictors.

\subsection*{Ethics statement}

The inD dataset consists of drone recordings of public road traffic, used under its provider's terms for non-commercial research use and not redistributed with this work. No personally identifying information is used and no human subjects were involved. The intended use of MaDE is post-hoc feasibility checking of predicted trajectories, and its guarantees are model-relative, holding against a learned completion model and a supplied anchor rather than against the true system. It should not be read as a safety certificate for a deployed system.

\subsection*{Reproducibility statement}

Section~\ref{sec:method} defines the MaDE cell and its training objectives, and Algorithm~\ref{alg:made-training} gives the two-phase training procedure. Section~\ref{sec:experiments:metrics} defines the evaluation metrics, and Table~\ref{tab:metrics} gives the simulated metrics in symbols. For the simulated experiment, Section~\ref{sec:experiments:e01} gives the perturbation and noise protocol, the baselines and the scoring of each method. Appendix~\ref{app:e01-systems} specifies the four simulated systems, their data splits and what each seed varies. For the real-data experiment, Section~\ref{sec:experiments:e05} describes the predictors, MaDE's training, the constraint set and the baselines. Appendix~\ref{app:e01-systems} gives the real-data constraint bounds, the learned wheelbase, the inD split and both displacement filters with the counts they leave, and specifies the three predictor families and their training. The rest of Appendix~\ref{app:solvers} gives the integrator, the inverse priors, the network architectures, the optimisers, the early-stopping policy and the baseline implementations, and tabulates the default hyperparameters. Appendix~\ref{app:solvers:optimisation} gives the epoch budget, batch size and validation schedule for the simulated systems. Appendices~\ref{app:e01-ablations}, \ref{app:prior-only} and~\ref{app:completion-only} define the MaDE ablations, the prior-only comparator and the completion-only variant. Appendix~\ref{app:runtime} gives the hardware, the training compute and the inference timing protocol and results.

\begin{ack}
We thank the anonymous reviewers and the area chair for their feedback, which shaped this version of the paper. Kevin Yu is supported by the Skempton Scholarship from the Department of Civil and Environmental Engineering at Imperial College London and by the Imperial-TUM Joint Academy of Doctoral Studies. Tao Guo is supported by the ATHLOS project (Project No. J2301), funded by the International Graduate School of Science and Engineering (IGSSE) of the Technical University of Munich. The authors declare no competing interests.
\end{ack}

\bibliographystyle{plainnat}
\bibliography{references}

\appendix

\section{Scope and Failure Modes}
\label{app:scope-failure}

By construction, MaDE returns $(x_t, u_{t-1})$ with $x_t = \mathcal{T}_\theta(x_{t-1}, u_{t-1}; \Delta t)$ up to solver tolerance: every iterate of the correction update in Eq.~\eqref{eq:made-correction-update} re-evaluates $\mathcal{T}_\theta$ at the current control, so the returned state is never edited independently of the dynamics. Likewise, if the corrector terminates at a control with $J(u_{t-1}) = 0$, the definition of $J$ in Eq.~\eqref{eq:made-correction-objective} forces $g(x_t, u_{t-1}) \leq 0$ componentwise. Both statements are model-relative and conditional: they hold relative to the learned $\mathcal{T}_\theta$ and the supplied anchor $x_{t-1}$, and make no claim about agreement with the true system, optimality of the correction, or feasibility of the anchor itself.

These properties are operative only within the following scope:
\begin{enumerate}
    \item \textbf{One-step locality.} Constraints are evaluated from a single transition $(x_t, u_{t-1})$. Trajectory-level constraints (e.g., total energy budgets, multi-step path constraints) are out of scope.
    \item \textbf{State sufficiency.} $x_t$ contains every variable needed to evaluate one-step dynamics and constraints. Hidden multi-step modes cannot be recovered by a Markov, time-invariant operator.
    \item \textbf{Parameter availability.} The static physical parameters of the known model are fixed at evaluation, supplied or fitted, and MaDE does not learn them online. On the simulated systems they are supplied per condition. On inD the wheelbase is the 2.7 m reference wheelbase plus a per-vehicle residual from a metadata encoder, which is fitted during training and frozen with the rest of MaDE (Appendix~\ref{app:e01-systems}). There the two properties hold relative to the learned $\mathcal{T}_\theta$ at that per-vehicle wheelbase and the supplied anchor $x_{t-1}$.
    \item \textbf{Known physics is meaningful.} The APHYNITY decomposition presumes that $f_{\mathrm{phys}}$ explains a substantial fraction of the dynamics. Otherwise $F_{a,\theta}$ carries the entire model and the minimum-norm penalty becomes a drag rather than a regulariser. Under the assumptions the APHYNITY decomposition requires, the learned residual is interpretable as the minimum-norm complement to the known physical family. Lifted from autonomous to controlled systems, this argument is best read as motivation rather than re-proof: it transfers to the lifted vector-field domain $X \times U \times P$ when the analogous projection-style assumptions hold for the chosen physics family.
    \item \textbf{The known model substitutes rather than omits.} On DB the presumption of item 4 must be read channel by channel, and it does not hold for the heading, lateral-velocity or yaw-rate derivatives. The known kinematic bicycle is embedded to act on the six-dimensional state, a pairing of a kinematic model with a learned residual that has been done before \citep{teng2025correcting}. It returns zero for the derivatives of lateral velocity and yaw rate, and reads neither variable when computing its other four derivatives. In the true system the heading rate equals the yaw rate exactly, and the yaw rate is an observed state component. The kinematic model instead computes a heading rate from longitudinal velocity and steering angle, and its position derivatives omit the lateral-velocity contribution the true system carries. $F_{a,\theta}$ must therefore cancel the kinematic heading-rate term and substitute the yaw rate, in addition to supplying the two absent derivatives. The minimum-norm penalty then acts against a residual that cannot be small on the affected channels. The minimum-norm interpretation of the residual still applies, because the APHYNITY decomposition places no condition distinguishing a prior that omits a term from one that substitutes for it.
    \item \textbf{Anchor exogeneity.} MaDE treats the supplied previous-state anchor $x_{t-1}$ as exogenous and never corrects it, so rollouts inherit anchor feasibility from whatever produces the anchor.
    \item \textbf{Inverse-model error on out-of-distribution proposals.} $\mathcal{I}_\phi$ is trained on feasible transitions, so a far-out-of-distribution upstream proposal $\tilde{x}_t$ can yield an out-of-distribution inferred control. Some forcing mismatch may then be absorbed by $F_{a,\theta}$, weakening the APHYNITY interpretation. Cycle consistency, alternating optimisation, and the inverse-residual penalty $\mathcal{R}_{\Delta\mathcal{I}}$ are mitigations, not removals.
    \item \textbf{Finite-budget non-termination.} The fixed-iteration corrector $\mathcal{C}_g$ is not guaranteed to reach $J(u) = 0$ within budget. When it stops short, the dynamics property above still holds for the returned transition, and the inequality property does not.
\end{enumerate}

Outside these conditions MaDE remains a well-defined algorithm, but its outputs lose their interpretation as projections onto a learned-dynamics-feasible set.

The evidence for robustness to upstream error comes from the experiments' perturbation protocol, not from a distribution-free analytic claim. The simulated experiments establish how the operator responds to a prescribed perturbation, not how it behaves under the error distribution of a deployed predictor. The architecture ablations were run in simulation only, so whether their ordering holds on real data is untested.

Anchor exogeneity and finite-budget non-termination (items 6 and 8) interact in a recursive rollout, where the anchor is MaDE's own previous output. An unresolved violation at one step then initialises the next step from an infeasible anchor. A set is controlled invariant when, from every state in it, some control inside the admissible box $[u_{\min}, u_{\max}]$ keeps the next state under $\mathcal{T}_\theta$ inside the set. If the anchor falls outside the largest controlled-invariant subset of $\{g \leq 0\}$, the one-step forward reachable set under $\mathcal{T}_\theta$ and that control box need not intersect $\{g \leq 0\}$. In that case no admissible control restores satisfaction of the inequalities within one step, and gradient descent on $J$ stalls at a non-zero minimum. MaDE never edits the anchor, so it has no mechanism for recovering from this state. Characterising the conditions under which recursive application keeps the anchor inside a controlled-invariant set is left to future work.
\section{Experimental Details}
\label{app:solvers}

\begin{algorithm}[h]
\caption{MaDE training (Phases 1 and 2)}
\label{alg:made-training}
\begin{algorithmic}[1]
\Require Feasible transitions $(x_{t-1}, x_t)$, control sampler $q(u)$, constraint map $g$, weights $\lambda_{\mathrm{inv}}, \lambda_a, \lambda_{\Delta\mathcal{I}}, \lambda_g$
\For{training iteration}
    \State Sample a minibatch of feasible transitions
    \State \textbf{Phase 1 I-side:} update $\mathcal{I}_\phi$ using $\mathcal{L}_{\mathrm{fwd}} + \lambda_{\mathrm{inv}}\mathcal{L}_{\mathrm{inv}} + \lambda_{\Delta\mathcal{I}} \mathcal{R}_{\Delta\mathcal{I}}$, with dynamics gradients frozen
    \State \textbf{Phase 1 T-side:} sample controls $u \sim q(u)$ and update $\mathcal{T}_\theta$ using $\mathcal{L}_{\mathrm{fwd}} + \lambda_a \mathcal{R}_a$, with inverse-model gradients frozen
    \If{inequality-aware training is enabled}
        \State Run $\mathcal{I}_\phi$, $\mathcal{T}_\theta$, and a fixed differentiable correction loop $\mathcal{C}_g$
        \State \textbf{Phase 2 I-side:} update $\mathcal{I}_\phi$ using $\mathcal{L}_{\mathrm{fwd}} + \lambda_{\mathrm{inv}}\mathcal{L}_{\mathrm{inv}} + \lambda_{\Delta\mathcal{I}} \mathcal{R}_{\Delta\mathcal{I}} + \lambda_g \mathcal{L}_{\mathrm{ineq}}$
        \State \textbf{Phase 2 T-side:} update $\mathcal{T}_\theta$ using $\mathcal{L}_{\mathrm{fwd}} + \lambda_a \mathcal{R}_a$, with inverse-model gradients frozen and no gradient from $\mathcal{L}_{\mathrm{ineq}}$
    \EndIf
\EndFor
\State \Return trained MaDE cell $\mathcal{M}_{\Theta,g}$
\end{algorithmic}
\end{algorithm}

Algorithm~\ref{alg:made-training} states the two-phase training procedure of Section~\ref{sec:method}.

\subsection{System Specifications}
\label{app:e01-systems}

\begin{table}[h]
\caption{Evaluation metrics for the simulated systems. A hat marks a returned state or the control a metric is scored on.}
\label{tab:metrics}
\centering
\begin{tabular}{ll}
\toprule
Metric & Definition \\
\midrule
Ineq.\ rate & Fraction of steps violating analytical bounds $g(\hat{x}_t,\hat{u}_{t-1}) \leq 0$ \\
Ineq.\ mag. & Mean $\|\mathrm{ReLU}(g(\hat{x}_t,\hat{u}_{t-1}))\|_2$ over violating steps \\
Dyn.-K & $\|\hat{x}_t-\mathcal{T}_{\mathrm{phys}}(\hat{x}_{t-1},\hat{u}_{t-1})\|_2$ \\
Dyn.-L & $\|\hat{x}_t-(\mathcal{T}_{\mathrm{phys}}+F_a)(\hat{x}_{t-1},\hat{u}_{t-1})\|_2$ \\
Dyn.-T & $\|\hat{x}_t-\mathcal{T}^{\mathrm{true}}(\hat{x}_{t-1},\hat{u}_{t-1})\|_2$ \\
Fid. & Mean $\|\hat{x}_t-x_t^{\mathrm{gt}}\|_2$ against the unperturbed simulator trajectory \\
\bottomrule
\end{tabular}
\end{table}

\paragraph{Double integrator (DI).} State $x = (x, y, \dot x, \dot y)$ and control $u = (a_x, a_y)$. The vector field is linear, with no heading, geometry or physics parameters. Inequalities are box bounds on position, velocity, and acceleration, augmented with a nonlinear speed-norm bound $\|(\dot x, \dot y)\|_2 \le v_{\max}$.

\paragraph{Unicycle (UNI).} State $x = (x, y, \theta, v)$ and control $u = (\delta, a)$, the heading rate and longitudinal acceleration. Position couples to heading through $\sin\theta$ and $\cos\theta$, with no vehicle geometry, no inertial coupling and no physics parameters. Inequalities are box bounds on position, $\theta$, $v$, $\delta$, and $a$.

\paragraph{Kinematic bicycle (KB).} State $x = (x, y, \theta, v)$ and control $u = (\delta, a)$, the steering angle and longitudinal acceleration. The single physics parameter is the wheelbase $L$. Inequalities are the box bounds of UNI.

\paragraph{Dynamic bicycle (DB).} State $x = (x, y, \theta, v_x, v_y, \dot\theta)$ and control $u = (\delta, a)$. The six physics parameters $(C_f, C_r, m, I_z, l_f, l_r)$ are the front and rear cornering stiffnesses, vehicle mass, yaw moment of inertia, and the front and rear wheelbase distances. Inequalities are box bounds on position, $\theta$, $v_x$, $v_y$, $\dot\theta$, $\delta$, and $a$. The controls that generate every DB trajectory are drawn uniformly from $\delta \in [-0.2, 0.2]$ rad and $a \in [-1.5, 1.5]$ metres per second squared, inside the inequality bounds of $\pm 0.5$ rad and $\pm 3.0$ metres per second squared. Initial states are drawn uniformly from the state bounds, narrowed on three channels to $v_x \geq 2.0$ metres per second, $v_y \in [-1.0, 1.0]$ metres per second, and $\dot\theta \in [-0.3, 0.3]$ rad per second. The Gaussian training augmentation empirically stabilised training in the underspecified DB condition, where MaDE's known forward and inverse model is the kinematic bicycle (Section~\ref{sec:experiments:e01}).

\paragraph{Simulated data and seeds.} The trajectories of each simulated condition are partitioned into $1024$ training, $128$ validation, and $128$ test trajectories. Each seed has its own parameter initialisation and minibatch order on the same generated trajectories. The bound-violation evaluation perturbation is identical across seeds, and on DB the observation noise and training augmentation of Section~\ref{sec:experiments:e01} are drawn afresh for each seed.

\paragraph{Real-data experiment (inD).} The real-data constraint set bounds heading, speed, steering angle, and acceleration. Speed is bounded from 0 to 22.0 metres per second (about 79 km/h), steering angle from $-0.5$ to $0.5$ rad (about $\pm 29^\circ$), and acceleration from $-8.0$ to $4.0$ metres per second squared. The position channels carry no bound, and the recorded positions do not support a reliable one. The simulated systems use the position bounds that generated their data, which are known exactly.

On inD, MaDE's known model is the kinematic bicycle at the reference wheelbase $L_{\mathrm{ref}} = 2.7$ m, and a metadata encoder adds a signed per-vehicle residual $\Delta L$, so that $L = L_{\mathrm{ref}} + \Delta L$. The encoder reads the vehicle's recorded length and width, two class indicators and a learned embedding of the recording location, and its output is neither bounded nor clamped. Across the 1,265 test-split vehicles the learned $L$ has a mean of 2.9238, 2.7879 and 2.7196 m for the models trained with seeds 0, 1 and 2, with population standard deviations of 0.0449, 0.0815 and 0.0530 m. For the same models it ranges from 2.8797 to 3.2265 m, from 2.6884 to 3.3163 m and from 2.6675 to 3.1035 m. The sign of $\Delta L$ depends on the seed: it is positive for all 1,265 vehicles under the seed-0 model, and negative for 3 vehicles under the seed-1 model and for 294 under the seed-2 model. The Pearson correlation of the learned $L$ with vehicle length is 0.987, 0.997 and 0.996 for the three models.

\paragraph{inD split.} Each of the 33 inD recordings is assigned whole to one split. Within each of the four intersections one recording is held out for validation and one for test, and the rest are used for training. That gives 25 training, 4 validation and 4 test recordings, with every intersection in every split. The assignment is fixed by hand, so no random seed enters it. Car, truck and bus tracks are kept, and pedestrian and bicycle tracks are dropped. Each track is downsampled from 25 Hz to 5 Hz, and a track shorter than 8 frames at 5 Hz is dropped, which leaves 5,798 training, 1,155 validation and 1,265 test tracks. Prediction windows pair a 10-step history with a 15-step horizon and start every 5 steps along each track.

\paragraph{Displacement filters.} The window-level filter excludes a prediction window when its net displacement over the 3 s horizon is at most 0.5 m. It keeps 33,329 of 289,075 training, 5,551 of 50,114 validation and 6,681 of 54,465 test windows, and the predictors are trained, validated and evaluated on those windows. The track-level filter excludes a track when its end-to-end displacement over its valid length is at most 0.5 m, and keeps every transition pair of a surviving track. A moving track that pauses therefore keeps those stationary transitions in MaDE's training, while an evaluation window inside that pause is removed. In both of MaDE's training phases the track-level filter keeps 315,288 of 1,566,934 training pairs and 56,813 of 274,778 validation pairs.

\paragraph{inD predictors.} The state of an inD vehicle is its position, heading and speed, $x = (x, y, \theta, v)$. Each of the 10 history states enters every predictor as five features: the position offset from the last observed state, the cosine and sine of the heading, and the speed, with the offset scaled and the speed standardised by training-split statistics. Each predictor also receives the vehicle's recorded length and width, two class indicators and a learned 4-dimensional embedding of its recording location, and conditions on them through a projection initialised at zero.

The recurrent model runs an LSTM cell of hidden size 64 over the history features, with its initial hidden and cell states set from the metadata by a ReLU MLP with one hidden layer of width 64. Its final hidden state feeds a ReLU MLP decoder with two hidden layers of width 128, which emits all 15 state deltas at once. The compact structured state-space (S6/Mamba-style) sequence model projects each history step to width 64 and applies two pre-norm residual blocks. Each block expands to an inner width of 128 and applies a depthwise causal convolution of kernel size 4 followed by a selective scan with state dimension 16. The scan's initial state is a linear function of the metadata. After a final layer normalisation, the last step's output feeds a decoder of the same shape as the recurrent model's. The transformer embeds each history step at width 64 with a sinusoidal positional encoding and applies two pre-norm encoder layers, each with 4 attention heads and a GELU feed-forward block of width 128. Its decoder holds 15 learned query vectors, one per horizon step and each offset by a linear function of the metadata, which cross-attend once to the encoded history. A ReLU MLP with two hidden layers of width 128 maps each attended query to its step's state delta.

Each predictor minimises the mean-squared error between its forecast and the recorded future states, with Adam at a constant learning rate of $10^{-3}$ on minibatches of 32 windows. Validation mean-squared error is computed after every epoch, and training stops once it has failed to improve for 10 epochs, after a burn-in of 5 epochs and under a ceiling of 200 epochs. The weights with the lowest validation error are kept, and the 15 runs stopped after 29 to 69 epochs. Each family is trained at seeds 0 to 4, and the seed sets the parameter initialisation and the minibatch order.

\subsection{ODE Integration}

$\mathcal{T}_\theta$ holds the control constant over each step, on the simulated systems and on inD alike. It is integrated with the embedded 2(1) Heun method (\texttt{diffrax.Heun}) under \texttt{diffrax.ConstantStepSize}, taking exactly one fixed step of size $\Delta t$ over $[0, \Delta t]$. The step is $\Delta t = 0.1$ on the four simulated systems and $\Delta t = 0.2$ s in the real-data experiment. Backpropagation uses \texttt{diffrax.RecursiveCheckpointAdjoint} for the standalone forward pass and \texttt{diffrax.DirectAdjoint} inside the corrector's inner violation-gradient solve, where reverse-mode differentiation through the nested correction loop requires it. All integration runs in \texttt{float64}. The ODE solvers are unstable at \texttt{float32}, with NaN gradients near the low-$v_x$ region of the dynamic bicycle.

Dyn.-K and Dyn.-T integrate the relevant pure-physics vector field, and Dyn.-L the augmented dynamics, under the same fixed-step Heun convention.

\subsection{Inverse Priors}
\label{app:solvers:inverse-priors}

The analytical control prior $\mathcal{I}_{\mathrm{known}}$ of Eq.~\eqref{eq:made-inverse-decomp} is implemented per system. It is a structural baseline rather than an exact inverse of the numerical ODE solve.

\paragraph{Double integrator.} Acceleration is recovered from the velocity finite difference $a = (\dot x_t - \dot x_{t-1}) / \Delta t$, which is exact under continuous-time double-integrator dynamics with zero-order-hold control.

\paragraph{Unicycle.} Heading rate and acceleration are recovered as finite differences $\delta = (\theta_t - \theta_{t-1}) / \Delta t$ and $a = (v_t - v_{t-1}) / \Delta t$, exact under zero-order-hold inputs.

\paragraph{Kinematic bicycle.} Heun integration of $\dot \theta = v \tan\delta / L$ under constant $(\delta, a)$ has the closed-form average velocity $v_{\mathrm{avg}} = \tfrac{1}{2}(v_{t-1} + v_t)$. This gives the Heun-exact inverse $\delta = \arctan\!\bigl(L \, \Delta\theta \,/\, (v_{\mathrm{avg}} \, \Delta t)\bigr)$ and $a = (v_t - v_{t-1}) / \Delta t$, with a numerical floor of $10^{-6}$ on $|v_{\mathrm{avg}}|$ to avoid the near-zero-velocity singularity. The real-data experiment uses the same inverse but sets the steering angle to zero wherever $|v_{\mathrm{avg}}|$ is below 0.5 m/s, leaving the acceleration unchanged. Jitter in the recorded heading would otherwise turn the near-zero-velocity singularity into spurious steering.

\paragraph{Dynamic bicycle.} The dynamic bicycle has no closed-form inverse under Heun integration, so its prior is a two-stage Newton scheme aligned with the training integrator. Stage A initialises $\delta$ from the kinematic prior, runs two unrolled Newton iterations on the algebraic yaw-rate finite-difference equation for $\delta$, and then closes the form for $a$ from the $v_x$ equation given the solved $\delta$. Stage B runs three outer Newton iterations on the two-dimensional residual
\begin{equation*}
r(\delta, a) \;=\; \bigl(\,\mathrm{Heun}(x_{t-1}, [\delta, a])_{[v_x]} - x_{t,[v_x]},\ \ \mathrm{Heun}(x_{t-1}, [\delta, a])_{[\dot\theta]} - x_{t,[\dot\theta]}\bigr),
\end{equation*}
whose two rows pick the $v_x$ and yaw-rate components, the two state dimensions most sensitive to $a$ and $\delta$ respectively. The $2 \times 2$ Jacobian is computed by forward-mode autodiff and inverted in closed form, with a determinant floor of $10^{-12}$ for gradient stability. At $\Delta t = 0.1$ the resulting prior matches the Heun simulator to roughly $10^{-12}$ on the tested manoeuvres. The underspecified-condition prior is the kinematic prior wrapped to the dynamic state vector, since the underspecified case is kinematic by design, and it is deliberately not promoted to a dynamic prior.

\paragraph{Parameter normalisation.} The dynamic-bicycle parameters $(C_f, C_r, m, I_z, l_f, l_r)$ span four orders of magnitude, so the parameter input to both $\Delta\mathcal{I}_\phi$ and the residual $F_a$ is normalised by per-system scales that give the MLPs $\mathcal{O}(1)$ inputs. The kinematic-bicycle wheelbase is normalised by $3$, and the dynamic-bicycle scales are $(2 \times 10^4,\ 2 \times 10^4,\ 1500,\ 3000,\ 2,\ 2)$. The known physics still receives the raw parameters.

\subsection{Network Architectures}

$\Delta \mathcal{I}_\phi$ is a two-hidden-layer MLP of width $256$ with ReLU activation, taking $(x_{t-1}, \tilde x_t)$ together with the physics parameters in normalised form (Appendix~\ref{app:solvers:inverse-priors}, ``Parameter normalisation''), and emitting a control increment of dimension $n_u$. Its final-layer weights and biases are scaled by zero at initialisation, so $\mathcal{I}_\phi$ starts equal to the analytical prior. The residual vector field $F_a$ is also a two-hidden-layer ReLU MLP of width $256$, taking $(X(\tau), u_{t-1})$ and the same normalised parameters and emitting a state-derivative increment. Its final layer is zero-scaled in the same way, so the augmented dynamics begin at the known physics. On the simulated systems the physics parameters are fixed inputs supplied at runtime. On inD the wheelbase comes from the metadata encoder of Appendix~\ref{app:e01-systems}, a two-hidden-layer ReLU MLP of width $64$ over four metadata columns and an $8$-dimensional location embedding. The encoder is updated with $F_a$ in the T-side step of Algorithm~\ref{alg:made-training}.

\subsection{Optimisation}
\label{app:solvers:optimisation}

\paragraph{Optimisers.} On the simulated systems runs use a maximum budget of $100$ epochs per phase at batch size $512$. In the real-data experiment MaDE uses a maximum budget of $150$ epochs per phase at batch size $128$. We use two Adam optimisers, one over the inverse-side parameters (the $\Delta\mathcal{I}_\phi$ MLP) and one over the dynamics-side parameters ($F_a$, and on inD also the wheelbase encoder). Each side uses a linear warmup of $100$ steps from $0$ to $10^{-3}$ followed by a constant rate, with a global gradient norm clip of $1.0$. The two sides alternate every $10$ steps unless noted otherwise.

\paragraph{Phase 1 cycle consistency.} Phase 1 combines $\mathcal{L}_{\mathrm{fwd}}$, $\mathcal{L}_{\mathrm{inv}}$, $\mathcal{R}_a$ and $\mathcal{R}_{\Delta\mathcal{I}}$ at the weights of Table~\ref{tab:defaults}. Inverse-consistency synthetic controls are drawn under a mixture strategy that interpolates from a uniform prior over the control box to the inverse-dynamics prediction across training, with mixing weight $\alpha = \mathrm{step} / \mathrm{total\_steps}$.

\paragraph{Phase 2 inequality-aware training.} Phase 2 adds $\mathcal{L}_{\mathrm{ineq}}$ at the weight $\lambda_g$ of Table~\ref{tab:defaults}, evaluated through a fixed-budget differentiable correction loop. The default DB result evaluates the Gaussian-augmented underspecified model directly under its native Gaussian regime, with no curriculum-specific training.

\paragraph{Corrector.} At training time the corrector takes $5$ inner gradient steps with step size $\alpha = 0.01$ and momentum $\beta = 0$, implemented as a \texttt{jax.lax.fori\_loop} whose fixed count keeps the loop differentiable with a static shape. At evaluation time it instead runs a \texttt{jax.lax.while\_loop} until $\max g(x, u) \le 10^{-6}$, capped at $50$ iterations. Both loops apply the projected, re-completing update of Eq.~\eqref{eq:made-correction-update}.

\paragraph{Proximity to the upstream proposal.} The violation objective $J(u)$ in Eq.~\eqref{eq:made-correction-objective} carries no term penalising deviation from the upstream proposal. Gradient descent therefore moves the control along the direction of steepest violation reduction in control space, without reference to the waypoint, heading or speed the predictor intended. The division of labour is intentional: proximity to the proposal is left to $\mathcal{T}_\theta$ and $\mathcal{I}_\phi$, and the $J(u)$ step is isolated to inequality violations alone. Two mechanisms realise that division: the initialisation at the inferred control $\hat{u}_{t-1}$ supplied by $\mathcal{I}_\phi$, and the finite iteration budget, which at evaluation time stops once the maximum violation falls below the tolerance. Neither mechanism bounds the displacement explicitly, and the only hard limit on the corrected control is the box $[u_{\min}, u_{\max}]$.

\paragraph{Gradient of the inequality loss after correction.} The Phase 2 inequality loss $\mathcal{L}_{\mathrm{ineq}}$ of Eq.~\eqref{eq:made-ineq-loss} is evaluated on the output of the unrolled correction loop rather than on the uncorrected proposal. Whenever the five training-time steps drive the candidate strictly inside the region where $g \leq 0$, the ReLU is identically zero and the term contributes no gradient on that sample. The term is therefore active only on the samples the corrector does not resolve within its training budget, and there it drives proposals toward controls the corrector can resolve within five steps. A resolved sample is still supervised by the Phase 1 losses, but this term supplies no gradient toward reducing the iterations spent on it. In the training of every reported MaDE model it is stopped out of the dynamics-side loss, so it updates $\mathcal{I}_\phi$ and never reaches the forward residual $F_a$.

\paragraph{Early stopping.} MaDE is trained with validation-driven early stopping within each training phase, under a patience window, a minimum delta, and a minimum-epoch burn-in. A phase also stops on either of two physics-motivated criteria, an exactness criterion and a saturation criterion. When early stopping fires, the best validation checkpoint for that phase is restored before training proceeds. On the simulated systems validation runs once per epoch, with a minimum-epoch burn-in of $5$, patience of $10$ validations, and minimum delta of $0$. The exactness criterion fires when forward consistency, inverse consistency, residual minimum-norm, and inverse-residual minimum-norm all fall below $10^{-6}$. The saturation criterion fires when forward and inverse consistency stay below $10^{-6}$ over a sliding window of $5$ validations while the two minimum-norm penalties stay stable to within $10^{-5}$.

\subsection{Baselines}
\label{app:solvers:baselines}
We reimplement both of FAB's training phases, a reconstruction phase on feasible data and an adversarial latent-structuring phase that fits a discriminator alongside the autoencoder. Hyperparameters follow the source's Safety Gym setting, with the latent dimension set to eight, the six free state dimensions and two controls that determine a successor.

The EKF/RTS smoother of the real-data experiment takes the whole forecast because a fixed-interval smoother applied one transition pair at a time degrades to a forward filter. It is tuned for each predictor family on the training-split forecasts of the same five predictor runs it corrects at evaluation. Its covariances are diagonal, and every diagonal is measured on the training split. The smoother's measurements are the predictor's forecast, so its measurement covariance $R$ is the predictor's per-channel mean squared error against ground truth. The process covariance $Q$ is one scale factor times a measured diagonal, whose state block is the per-channel mean squared one-step residual of the known model on ground-truth trajectories. Its control block is the per-channel mean squared one-step increment of the controls that the known model's analytic inverse recovers from those trajectories. A common rescaling of the initial covariance $P_0$, $Q$ and $R$ leaves the Kalman gain and the estimate unchanged, so the estimate depends only on the ratio between the covariances. Throughout the sweep $R$ keeps its measured value and the state block of $P_0$ is pinned equal to it, so with that invariance a single factor on $Q$ suffices to sweep the ratio. The control block of $P_0$ lies outside that invariance: it is held at the training-split variance of the recovered controls and is not tuned. The ADE criterion and the residual criterion of Section~\ref{sec:experiments:e05} select the same scale factor in the recurrent and transformer families and factors half a decade apart in the state-space family.

Each real-data baseline is deterministic and has no MaDE seed, so each aggregates 15 cells, three predictor families crossed with five predictor seeds. Where the two criteria select the same factor, the two smoothers give identical results, so in the recurrent and transformer families the residual-tuned entries of Table~\ref{tab:e05-ind} are also those of the accuracy-tuned smoother. In the state-space family the accuracy-tuned smoother has an ADE of $0.7063 \pm 0.0180$ m, an FDE of $1.7723 \pm 0.0286$ m and a Dyn.-K of $0.0291 \pm 0.0007$, each the mean and population standard deviation over the 5 cells of that family. Its inequality components are in Appendix~\ref{app:ineq-breakdown}.

\subsection{Default Hyperparameters}
\label{app:defaults}

\begin{table}[h]
\caption{Default MaDE hyperparameters.}
\label{tab:defaults}
\centering
\begin{tabular}{ll}
\toprule
Setting & Value \\
\midrule
Corrector iterations, training & 5, fixed \\
Corrector iterations, evaluation & adaptive, tolerance $10^{-6}$, cap 50 \\
Corrector step size on $u$ & $0.01$ \\
$\lambda$ inverse consistency & $1.0$ \\
$\lambda$ inequality (Phase 2) & $1.0$ \\
$\lambda$ minimum-norm on $F_a$ & $0.01$ \\
$\lambda$ minimum-norm on $\Delta\mathcal{I}_\phi$ & $0.01$ \\
Gradient clipping, both parameter groups & $1.0$ \\
Warm-up steps & $100$ \\
Loss early-stopping tolerance & $10^{-6}$ \\
\bottomrule
\end{tabular}
\end{table}

This is one setting rather than a per-system search, and it is identical across all four simulated systems and every ablation. The evaluation cap bounds worst-case compute, but it becomes an accuracy setting on any proposal that reaches it first. In the real-data experiment 0.964\% of corrector calls stop at the cap (Appendix~\ref{app:runtime}). The fully specified systems are the diagnostic for the minimum-norm weights, because the decomposition predicts $F_a \to 0$ wherever the known physics already explains the data.
\section{Additional Simulated Results}
\label{app:sim-results}

\subsection{MaDE Ablations}
\label{app:e01-ablations}

Each ablation shares the train, validation and test splits of the headline MaDE run. The learned ablations are trained under the same two-phase protocol and a comparable optimisation budget unless stated otherwise. \textbf{MaDE w/o $F_a$} sets the residual identically to zero, so the dynamics reduce to the known physics. The APHYNITY decomposition predicts $F_a \to 0$ in the fully specified conditions, where this variant should therefore match full MaDE. On the underspecified DB its gap to full MaDE quantifies the missing tyre dynamics. \textbf{MaDE w/o $\mathcal{C}$} keeps the augmented dynamics and skips inequality correction, which isolates the corrector's contribution from the quality of the dynamics. \textbf{MaDE fixed-$\mathcal{I}$} disables $\Delta\mathcal{I}_\phi$ and uses only the analytical control prior for the inverse step. \textbf{MaDE supervised-$\mathcal{I}$} trains $\Delta\mathcal{I}_\phi$ against simulator ground-truth controls rather than through cycle consistency, so it is available only in simulation. It is a supervised operating point rather than a subtractive ablation, since it replaces cycle consistency instead of removing a component, and so does not speak to whether a component is needed.

On DI, UNI and KB, Dyn.-T scores MaDE, its variants and the three baselines on controls recovered through the inverse of the true model, which on these systems is the inverse of the known model. MaDE's Dyn.-L is below $10^{-15}$ on these three systems, as it is on DB. On every system a baseline's Dyn.-L takes the controls that MaDE's inverse $\mathcal{I}_\phi$ recovers from the baseline's returned states, and MaDE's augmented dynamics $\mathcal{T}_\theta$ gives the residual. The baselines' Dyn.-L on the three systems, which no table reports, is $0.2397$ to $0.8547$, and each of these nine means differs from the same baseline's Dyn.-K by at most $1.5 \times 10^{-5}$.

MaDE w/o $F_a$ has a known-physics residual of zero by definition rather than by learning. On DB its true-dynamics residual is below that of full MaDE in each of the five seeds. This reverses the ordering the decomposition predicts for the underspecified condition, where full MaDE should have the lower residual. MaDE fixed-$\mathcal{I}$ is lower on that residual than both, and below full MaDE in each of the five seeds. The predicted ordering appears in fidelity instead, where full MaDE has the lowest error of the five variants on DB (Table~\ref{tab:e01-ablations}).

MaDE w/o $\mathcal{C}$ and MaDE supervised-$\mathcal{I}$, the two variants that remove or replace a component shaping the optimisation, are the two that produce unstable seeds on the reported metrics, both on DB. MaDE w/o $\mathcal{C}$ has one seed at $1.0241$ on Dyn.-K against $0.0493$ to $0.2323$ for the other four, and the same seed is at $14.0704$ on Dyn.-T against $0.4489$ to $0.9623$. MaDE supervised-$\mathcal{I}$ has one seed at $1.4487$ on Dyn.-K against $0.0448$ to $0.6121$ for the other four. It has two seeds at $11.4278$ and $3.6965$ on Ineq.\ mag.\ against $0.1504$ to $0.3312$ for the other three. The same two seeds are at $84.8408$ and $13.9413$ on Dyn.-T against $0.3139$ to $0.3363$. MaDE w/o $F_a$ and MaDE fixed-$\mathcal{I}$, which disable the two components that add capacity, show no such seed on any condition or reported metric.

\begin{table}[h]
\caption{MaDE ablations on the four simulated systems, as the mean $\pm$ the population standard deviation over five seeds. Lower is better for every metric. Within each metric row the best mean is set in \textbf{bold} and the second-best \underline{underlined}. A dagger marks an entry with at least one seed whose value exceeds $1.0$ in the metric's own units and three times the median of the other four seeds.}
\label{tab:e01-ablations}
\centering
\footnotesize
\setlength{\tabcolsep}{2pt}
\renewcommand{\arraystretch}{0.88}
\begin{adjustbox}{max width=\linewidth}\begin{tabular}{llccccc}
\toprule
 & & \multicolumn{4}{c}{Ablations} & \multicolumn{1}{c}{Reference} \\
\cmidrule(lr){3-6}\cmidrule(lr){7-7}
Cond. & Metric & w/o $F_a$ & w/o $\mathcal{C}$ & fixed-$\mathcal{I}$ & sup.-$\mathcal{I}$ & MaDE \\
\midrule
        DI & Ineq. rate & $\mathbf{0.2034\,{\scriptscriptstyle\pm}\,0.0000}$ & \underline{$0.2927\,{\scriptscriptstyle\pm}\,0.0000$} & $\mathbf{0.2034\,{\scriptscriptstyle\pm}\,0.0000}$ & $\mathbf{0.2034\,{\scriptscriptstyle\pm}\,0.0000}$ & $\mathbf{0.2034\,{\scriptscriptstyle\pm}\,0.0000}$ \\
         & Ineq. mag. & \underline{$0.2428\,{\scriptscriptstyle\pm}\,0.0000$} & $8.4198\,{\scriptscriptstyle\pm}\,0.0000$ & $0.2428\,{\scriptscriptstyle\pm}\,0.0000$ & $\mathbf{0.2428\,{\scriptscriptstyle\pm}\,0.0000}$ & $0.2428\,{\scriptscriptstyle\pm}\,0.0000$ \\
         & Dyn.-K & $\mathbf{0.0000\,{\scriptscriptstyle\pm}\,0.0000}$ & $0.0000\,{\scriptscriptstyle\pm}\,0.0000$ & $0.0000\,{\scriptscriptstyle\pm}\,0.0000$ & $0.0000\,{\scriptscriptstyle\pm}\,0.0000$ & \underline{$0.0000\,{\scriptscriptstyle\pm}\,0.0000$} \\
         & Dyn.-L & $\mathbf{0.0000\,{\scriptscriptstyle\pm}\,0.0000}$ & $0.0000\,{\scriptscriptstyle\pm}\,0.0000$ & \underline{$0.0000\,{\scriptscriptstyle\pm}\,0.0000$} & $0.0000\,{\scriptscriptstyle\pm}\,0.0000$ & $0.0000\,{\scriptscriptstyle\pm}\,0.0000$ \\
         & Dyn.-T & $\mathbf{0.0000\,{\scriptscriptstyle\pm}\,0.0000}$ & $0.0000\,{\scriptscriptstyle\pm}\,0.0000$ & $0.0000\,{\scriptscriptstyle\pm}\,0.0000$ & $0.0000\,{\scriptscriptstyle\pm}\,0.0000$ & \underline{$0.0000\,{\scriptscriptstyle\pm}\,0.0000$} \\
         & Fid. & $\mathbf{2.2792\,{\scriptscriptstyle\pm}\,0.0000}$ & $2.5513\,{\scriptscriptstyle\pm}\,0.0000$ & \underline{$2.2792\,{\scriptscriptstyle\pm}\,0.0000$} & $2.2792\,{\scriptscriptstyle\pm}\,0.0000$ & $2.2792\,{\scriptscriptstyle\pm}\,0.0000$ \\
        \\[-0.4ex]
        UNI & Ineq. rate & $\mathbf{0.1042\,{\scriptscriptstyle\pm}\,0.0000}$ & \underline{$0.1754\,{\scriptscriptstyle\pm}\,0.0001$} & $\mathbf{0.1042\,{\scriptscriptstyle\pm}\,0.0000}$ & $\mathbf{0.1042\,{\scriptscriptstyle\pm}\,0.0000}$ & $\mathbf{0.1042\,{\scriptscriptstyle\pm}\,0.0000}$ \\
         & Ineq. mag. & $\mathbf{0.0537\,{\scriptscriptstyle\pm}\,0.0000}$ & $3.8501\,{\scriptscriptstyle\pm}\,0.0037$ & $0.0538\,{\scriptscriptstyle\pm}\,0.0000$ & \underline{$0.0538\,{\scriptscriptstyle\pm}\,0.0000$} & $0.0538\,{\scriptscriptstyle\pm}\,0.0000$ \\
         & Dyn.-K & $\mathbf{0.0000\,{\scriptscriptstyle\pm}\,0.0000}$ & $0.0000\,{\scriptscriptstyle\pm}\,0.0000$ & \underline{$0.0000\,{\scriptscriptstyle\pm}\,0.0000$} & $0.0000\,{\scriptscriptstyle\pm}\,0.0000$ & $0.0000\,{\scriptscriptstyle\pm}\,0.0000$ \\
         & Dyn.-L & $\mathbf{0.0000\,{\scriptscriptstyle\pm}\,0.0000}$ & $0.0000\,{\scriptscriptstyle\pm}\,0.0000$ & \underline{$0.0000\,{\scriptscriptstyle\pm}\,0.0000$} & $0.0000\,{\scriptscriptstyle\pm}\,0.0000$ & $0.0000\,{\scriptscriptstyle\pm}\,0.0000$ \\
         & Dyn.-T & $\mathbf{0.0000\,{\scriptscriptstyle\pm}\,0.0000}$ & $0.0000\,{\scriptscriptstyle\pm}\,0.0000$ & \underline{$0.0000\,{\scriptscriptstyle\pm}\,0.0000$} & $0.0000\,{\scriptscriptstyle\pm}\,0.0000$ & $0.0000\,{\scriptscriptstyle\pm}\,0.0000$ \\
         & Fid. & $1.4126\,{\scriptscriptstyle\pm}\,0.0000$ & $1.4940\,{\scriptscriptstyle\pm}\,0.0000$ & $\mathbf{1.4126\,{\scriptscriptstyle\pm}\,0.0000}$ & $1.4126\,{\scriptscriptstyle\pm}\,0.0000$ & \underline{$1.4126\,{\scriptscriptstyle\pm}\,0.0000$} \\
        \\[-0.4ex]
        KB & Ineq. rate & $\mathbf{0.1965\,{\scriptscriptstyle\pm}\,0.0000}$ & $0.4302\,{\scriptscriptstyle\pm}\,0.0002$ & $\mathbf{0.1965\,{\scriptscriptstyle\pm}\,0.0000}$ & \underline{$0.1966\,{\scriptscriptstyle\pm}\,0.0001$} & $\mathbf{0.1965\,{\scriptscriptstyle\pm}\,0.0000}$ \\
         & Ineq. mag. & $\mathbf{0.1428\,{\scriptscriptstyle\pm}\,0.0000}$ & $4.2880\,{\scriptscriptstyle\pm}\,0.0007$ & $0.1430\,{\scriptscriptstyle\pm}\,0.0001$ & \underline{$0.1429\,{\scriptscriptstyle\pm}\,0.0001$} & $0.1431\,{\scriptscriptstyle\pm}\,0.0001$ \\
         & Dyn.-K & $\mathbf{0.0000\,{\scriptscriptstyle\pm}\,0.0000}$ & $0.0003\,{\scriptscriptstyle\pm}\,0.0001$ & $0.0003\,{\scriptscriptstyle\pm}\,0.0000$ & \underline{$0.0003\,{\scriptscriptstyle\pm}\,0.0001$} & $0.0003\,{\scriptscriptstyle\pm}\,0.0000$ \\
         & Dyn.-L & $\mathbf{0.0000\,{\scriptscriptstyle\pm}\,0.0000}$ & $0.0000\,{\scriptscriptstyle\pm}\,0.0000$ & \underline{$0.0000\,{\scriptscriptstyle\pm}\,0.0000$} & $0.0000\,{\scriptscriptstyle\pm}\,0.0000$ & $0.0000\,{\scriptscriptstyle\pm}\,0.0000$ \\
         & Dyn.-T & $\mathbf{0.0000\,{\scriptscriptstyle\pm}\,0.0000}$ & $0.0003\,{\scriptscriptstyle\pm}\,0.0001$ & $0.0003\,{\scriptscriptstyle\pm}\,0.0000$ & \underline{$0.0003\,{\scriptscriptstyle\pm}\,0.0001$} & $0.0003\,{\scriptscriptstyle\pm}\,0.0000$ \\
         & Fid. & $3.6796\,{\scriptscriptstyle\pm}\,0.0000$ & $3.9589\,{\scriptscriptstyle\pm}\,0.0005$ & \underline{$3.6794\,{\scriptscriptstyle\pm}\,0.0002$} & $\mathbf{3.6792\,{\scriptscriptstyle\pm}\,0.0001}$ & $3.6797\,{\scriptscriptstyle\pm}\,0.0006$ \\
        \\[-0.4ex]
        DB & Ineq. rate & $\mathbf{0.1279\,{\scriptscriptstyle\pm}\,0.0015}$ & $0.4862\,{\scriptscriptstyle\pm}\,0.1427$ & \underline{$0.1279\,{\scriptscriptstyle\pm}\,0.0010$} & $0.2088\,{\scriptscriptstyle\pm}\,0.1125$ & $0.1384\,{\scriptscriptstyle\pm}\,0.0055$ \\
         & Ineq. mag. & $\mathbf{0.1657\,{\scriptscriptstyle\pm}\,0.0014}$ & $4.8199\,{\scriptscriptstyle\pm}\,1.1599$ & \underline{$0.1661\,{\scriptscriptstyle\pm}\,0.0046$} & $3.1540\,{\scriptscriptstyle\pm}\,4.3515$$^\dagger$ & $0.2607\,{\scriptscriptstyle\pm}\,0.0392$  \\
         & Dyn.-K & $\mathbf{0.0000\,{\scriptscriptstyle\pm}\,0.0000}$ & $0.2873\,{\scriptscriptstyle\pm}\,0.3744$$^\dagger$ & \underline{$0.0494\,{\scriptscriptstyle\pm}\,0.0016$} & $0.4467\,{\scriptscriptstyle\pm}\,0.5451$$^\dagger$ & $0.0733\,{\scriptscriptstyle\pm}\,0.0326$  \\
         & Dyn.-L & $\mathbf{0.0000\,{\scriptscriptstyle\pm}\,0.0000}$ & $0.0000\,{\scriptscriptstyle\pm}\,0.0000$ & \underline{$0.0000\,{\scriptscriptstyle\pm}\,0.0000$} & $0.0000\,{\scriptscriptstyle\pm}\,0.0000$ & $0.0000\,{\scriptscriptstyle\pm}\,0.0000$ \\
         & Dyn.-T & \underline{$0.3049\,{\scriptscriptstyle\pm}\,0.0024$} & $3.3120\,{\scriptscriptstyle\pm}\,5.3822$$^\dagger$ & $\mathbf{0.1933\,{\scriptscriptstyle\pm}\,0.0031}$ & $19.9510\,{\scriptscriptstyle\pm}\,32.8707$$^\dagger$ & $0.3736\,{\scriptscriptstyle\pm}\,0.0647$  \\
         & Fid. & $5.4978\,{\scriptscriptstyle\pm}\,0.0153$ & $6.6773\,{\scriptscriptstyle\pm}\,1.7626$ & \underline{$5.4608\,{\scriptscriptstyle\pm}\,0.0238$} & $7.9124\,{\scriptscriptstyle\pm}\,3.8902$ & $\mathbf{5.2021\,{\scriptscriptstyle\pm}\,0.2740}$ \\
\bottomrule
\end{tabular}
\end{adjustbox}%
\end{table}

\subsection{Prior-Only Comparator}
\label{app:prior-only}

The prior-only comparator is the known physics, the analytic inverse control prior and the same corrector, with no trainable parameters, 12 bookkeeping floats against 139,796 for full MaDE. Run standalone on the underspecified dynamic bicycle, it beats full MaDE on both inequality metrics. It also has the lower true-dynamics residual, $0.3030 \pm 0.0007$ against $0.3736 \pm 0.0647$, each the mean and population standard deviation over five evaluations. The comparator's five evaluations differ only in the observation-noise draw they are scored on, while those of full MaDE also differ in its training seed. The comparator is below full MaDE on that residual in each of the five paired evaluations, and full MaDE keeps the best fidelity. This configuration was run after the submission and is not a column in the ablation table.

\subsection{Control Recovery}
\label{app:ctrl-recovery}

Table~\ref{tab:ctrl-recovery} compares the true controls with those recovered by the known inverse, which is the analytical control prior alone, and by MaDE and supervised-$\mathcal{I}$. On DI and UNI, where the supplied model is exact, every variant recovers the true controls to the discretisation floor, at or below $2.2 \times 10^{-5}$ nRMSE in every seed. On KB the analytic control prior recovers steering to $5.0 \times 10^{-15}$ nRMSE, and the learned inverse recovers it to $0.0010 \pm 0.0005$. On the underspecified dynamic bicycle no variant recovers the true controls. A negative acceleration bias remains in all three columns, the known inverse included, so it belongs to the setting rather than the learned model. These results are why Section~\ref{sec:method:inverse} describes the inferred control as model-relative.

\begin{table}[h]
\caption{Control recovery on the four simulated systems. nRMSE is normalised by the standard deviation of the true control on each channel. The known inverse is deterministic and has no spread, and the other two columns are the mean $\pm$ the population standard deviation across seeds. Lower nRMSE is better.}
\label{tab:ctrl-recovery}
\centering
\small
\setlength{\tabcolsep}{3pt}
\begin{tabular}{llccc}
\toprule
System & Channel & known inverse & MaDE & sup.-$\mathcal{I}$ \\
\midrule
DI (exact) & $a_x$ (nRMSE) & $7.90\times 10^{-16}$ & $2.36\times 10^{-6}\,{\scriptscriptstyle\pm}\,1.14\times 10^{-6}$ & $6.23\times 10^{-6}\,{\scriptscriptstyle\pm}\,6.39\times 10^{-6}$ \\
DI (exact) & $a_y$ (nRMSE) & $9.54\times 10^{-16}$ & $2.44\times 10^{-6}\,{\scriptscriptstyle\pm}\,1.01\times 10^{-6}$ & $2.36\times 10^{-6}\,{\scriptscriptstyle\pm}\,5.24\times 10^{-7}$ \\
UNI (exact) & $\delta$ (nRMSE) & $1.34\times 10^{-15}$ & $9.97\times 10^{-6}\,{\scriptscriptstyle\pm}\,2.81\times 10^{-6}$ & $1.12\times 10^{-5}\,{\scriptscriptstyle\pm}\,5.45\times 10^{-6}$ \\
UNI (exact) & $a$ (nRMSE) & $1.17\times 10^{-15}$ & $6.99\times 10^{-6}\,{\scriptscriptstyle\pm}\,1.63\times 10^{-6}$ & $7.44\times 10^{-6}\,{\scriptscriptstyle\pm}\,1.34\times 10^{-6}$ \\
KB (exact) & $\delta$ (nRMSE) & $5.04\times 10^{-15}$ & $1.01\times 10^{-3}\,{\scriptscriptstyle\pm}\,4.95\times 10^{-4}$ & $6.51\times 10^{-4}\,{\scriptscriptstyle\pm}\,3.20\times 10^{-4}$ \\
DB (underspec.) & $\delta$ (nRMSE) & $0.9200$ & $2.2256\,{\scriptscriptstyle\pm}\,0.3271$ & $2.0045\,{\scriptscriptstyle\pm}\,0.3555$ \\
DB (underspec.) & $a$ (nRMSE) & $0.2602$ & $0.2582\,{\scriptscriptstyle\pm}\,0.0051$ & $0.2586\,{\scriptscriptstyle\pm}\,0.0027$ \\
DB (underspec.) & $a$ (bias) & $-0.1594$ & $-0.1569\,{\scriptscriptstyle\pm}\,0.0058$ & $-0.1575\,{\scriptscriptstyle\pm}\,0.0033$ \\
\bottomrule
\end{tabular}
\end{table}

\subsection{Gradient-Depth Probe}
\label{app:grad-depth}

\begin{table}[h]
\caption{Gradient norms at five depths of the recursive corrector. Each row is the exact Phase-2 training gradient recomputed at that depth on the trained real-data model, over one fixed batch of 64 training transition pairs. Each norm is taken over one operator's own parameter subtree, before clipping.}
\label{tab:grad-depth}
\centering
\begin{tabular}{lcc}
\toprule
Correction depth & Norm on $\mathcal{I}_\phi$ & Norm on $\mathcal{T}_\theta$ \\
\midrule
0 & $12.0192$ & $0.079757$ \\
1 & $11.5167$ & $0.079757$ \\
2 & $11.5160$ & $0.079757$ \\
4 & $11.5148$ & $0.079757$ \\
8 & $11.5123$ & $0.079757$ \\
\bottomrule
\end{tabular}
\end{table}

The gradient norm on $\mathcal{I}_\phi$ neither diverges nor collapses across depths. It falls by $4.18\%$ from depth 0, where no correction is applied, to depth 1. From depth 1 it falls at each further probed depth, ending $0.038\%$ lower at depth 8. Averaged over each interval between probed depths from depth 1 onward, one added iterate lowers the norm on $\mathcal{I}_\phi$ by $6.280 \times 10^{-4}$ to $6.307 \times 10^{-4}$. The decrement is smallest over the deepest interval, from depth 4 to depth 8. The probed checkpoint was trained at corrector depth 5 on the 315,288 training pairs that survive the track-level filter, while the probe batch is drawn from all 1,566,934.

The norm on $\mathcal{T}_\theta$ is the same at every depth, and no statement about conditioning rests on it. The probed model was trained with the inequality term cut out of the dynamics-side gradient. The remaining dynamics-side terms contain no corrector, so the gradient reaching $\mathcal{T}_\theta$ cannot depend on the correction depth.
\section{Additional Real-Data Results}
\label{app:real-results}

\subsection{Inequality Components}
\label{app:ineq-breakdown}

Tables~\ref{tab:ineq-breakdown-ind-rate} and~\ref{tab:ineq-breakdown-ind-mag} split the real-data inequality violation rate and magnitude into a state-bound and a control-bound component. The inequality vector stacks the upper and lower state bounds and then the upper and lower control bounds, and each component is computed on the entries that belong to its own bounds. The state bounds are those on the vehicle state, including the speed bounds of Section~\ref{sec:experiments:e05}, and the control bounds are the steering-angle and acceleration bounds.

The combined rate is the fraction of steps that violate at least one state bound or control bound, counting a step that violates both once. It therefore lies between the larger of the two components and their sum, and falls short of the sum by the fraction of steps that violate both.

Each magnitude is the mean violation norm over the steps that violate the bounds it covers, averaged over windows. On a single step the combined norm is the Euclidean norm of the two component norms, because the state-bound and control-bound entries are disjoint parts of one violation vector. Averaging does not preserve that relation, since a mean of norms is not the norm of the means. In 104 of the 105 cells of the real-data experiment the combined mean lies between the Euclidean norm of the two component means and their sum. In the remaining cell the combined mean is 0.0030689, below its state component of 0.0030712, with a control component of $1.0 \times 10^{-10}$. The two bounds coincide where one component is zero, and there the combined mean equals both.

\begin{table}[h]
\caption{Inequality violation rate on the real-data experiment, by component. The rows are those of Table~\ref{tab:e05-ind} plus the accuracy-tuned smoother, and (acc.) and (res.) mark the accuracy-tuned and residual-tuned smoothers. Entries are the mean $\pm$ the population standard deviation over cells, 5 per predictor family for each baseline and 15 for MaDE. Lower is better.}
\label{tab:ineq-breakdown-ind-rate}
\centering
\begin{adjustbox}{max width=\linewidth}{\setlength{\tabcolsep}{3pt}
\begin{tabular}{llccc}
\toprule
Predictor & Row & state & control & combined \\
\midrule
\multirow{5}{*}{Recurrent} & raw & $0.0108\,{\scriptscriptstyle\pm}\,0.0034$ & $0.0483\,{\scriptscriptstyle\pm}\,0.0090$ & $0.0572\,{\scriptscriptstyle\pm}\,0.0093$ \\
 & clamp & $0.0000\,{\scriptscriptstyle\pm}\,0.0000$ & $0.0468\,{\scriptscriptstyle\pm}\,0.0091$ & $0.0468\,{\scriptscriptstyle\pm}\,0.0091$ \\
 & smoother (acc.) & $0.0250\,{\scriptscriptstyle\pm}\,0.0089$ & $0.0790\,{\scriptscriptstyle\pm}\,0.0071$ & $0.0855\,{\scriptscriptstyle\pm}\,0.0093$ \\
 & smoother (res.) & $0.0250\,{\scriptscriptstyle\pm}\,0.0089$ & $0.0790\,{\scriptscriptstyle\pm}\,0.0071$ & $0.0855\,{\scriptscriptstyle\pm}\,0.0093$ \\
 & MaDE & $0.0114\,{\scriptscriptstyle\pm}\,0.0059$ & $0.0000\,{\scriptscriptstyle\pm}\,0.0000$ & $0.0114\,{\scriptscriptstyle\pm}\,0.0059$ \\
\midrule
\multirow{5}{*}{State-space} & raw & $0.0110\,{\scriptscriptstyle\pm}\,0.0013$ & $0.0504\,{\scriptscriptstyle\pm}\,0.0067$ & $0.0583\,{\scriptscriptstyle\pm}\,0.0057$ \\
 & clamp & $0.0000\,{\scriptscriptstyle\pm}\,0.0000$ & $0.0478\,{\scriptscriptstyle\pm}\,0.0066$ & $0.0478\,{\scriptscriptstyle\pm}\,0.0066$ \\
 & smoother (acc.) & $0.0279\,{\scriptscriptstyle\pm}\,0.0028$ & $0.0779\,{\scriptscriptstyle\pm}\,0.0055$ & $0.0859\,{\scriptscriptstyle\pm}\,0.0050$ \\
 & smoother (res.) & $0.0271\,{\scriptscriptstyle\pm}\,0.0029$ & $0.0769\,{\scriptscriptstyle\pm}\,0.0055$ & $0.0849\,{\scriptscriptstyle\pm}\,0.0054$ \\
 & MaDE & $0.0111\,{\scriptscriptstyle\pm}\,0.0053$ & $0.0000\,{\scriptscriptstyle\pm}\,0.0000$ & $0.0111\,{\scriptscriptstyle\pm}\,0.0053$ \\
\midrule
\multirow{5}{*}{Transformer} & raw & $0.0072\,{\scriptscriptstyle\pm}\,0.0017$ & $0.0307\,{\scriptscriptstyle\pm}\,0.0083$ & $0.0358\,{\scriptscriptstyle\pm}\,0.0096$ \\
 & clamp & $0.0000\,{\scriptscriptstyle\pm}\,0.0000$ & $0.0288\,{\scriptscriptstyle\pm}\,0.0080$ & $0.0288\,{\scriptscriptstyle\pm}\,0.0080$ \\
 & smoother (acc.) & $0.0158\,{\scriptscriptstyle\pm}\,0.0024$ & $0.0578\,{\scriptscriptstyle\pm}\,0.0144$ & $0.0648\,{\scriptscriptstyle\pm}\,0.0134$ \\
 & smoother (res.) & $0.0158\,{\scriptscriptstyle\pm}\,0.0024$ & $0.0578\,{\scriptscriptstyle\pm}\,0.0144$ & $0.0648\,{\scriptscriptstyle\pm}\,0.0134$ \\
 & MaDE & $0.0065\,{\scriptscriptstyle\pm}\,0.0047$ & $0.0000\,{\scriptscriptstyle\pm}\,0.0000$ & $0.0065\,{\scriptscriptstyle\pm}\,0.0047$ \\
\bottomrule
\end{tabular}
}
\end{adjustbox}%
\end{table}

\begin{table}[h]
\caption{Inequality violation magnitude on the real-data experiment, by component, with rows and entries as in Table~\ref{tab:ineq-breakdown-ind-rate}. Lower is better.}
\label{tab:ineq-breakdown-ind-mag}
\centering
\begin{adjustbox}{max width=\linewidth}{\setlength{\tabcolsep}{3pt}
\begin{tabular}{llccc}
\toprule
Predictor & Row & state & control & combined \\
\midrule
\multirow{5}{*}{Recurrent} & raw & $0.0057\,{\scriptscriptstyle\pm}\,0.0018$ & $0.0317\,{\scriptscriptstyle\pm}\,0.0057$ & $0.0337\,{\scriptscriptstyle\pm}\,0.0048$ \\
 & clamp & $0.0000\,{\scriptscriptstyle\pm}\,0.0000$ & $0.0309\,{\scriptscriptstyle\pm}\,0.0058$ & $0.0309\,{\scriptscriptstyle\pm}\,0.0058$ \\
 & smoother (acc.) & $0.0178\,{\scriptscriptstyle\pm}\,0.0052$ & $0.0539\,{\scriptscriptstyle\pm}\,0.0077$ & $0.0650\,{\scriptscriptstyle\pm}\,0.0108$ \\
 & smoother (res.) & $0.0178\,{\scriptscriptstyle\pm}\,0.0052$ & $0.0539\,{\scriptscriptstyle\pm}\,0.0077$ & $0.0650\,{\scriptscriptstyle\pm}\,0.0108$ \\
 & MaDE & $0.0034\,{\scriptscriptstyle\pm}\,0.0021$ & $0.0000\,{\scriptscriptstyle\pm}\,0.0000$ & $0.0034\,{\scriptscriptstyle\pm}\,0.0021$ \\
\midrule
\multirow{5}{*}{State-space} & raw & $0.0079\,{\scriptscriptstyle\pm}\,0.0013$ & $0.0371\,{\scriptscriptstyle\pm}\,0.0028$ & $0.0397\,{\scriptscriptstyle\pm}\,0.0030$ \\
 & clamp & $0.0000\,{\scriptscriptstyle\pm}\,0.0000$ & $0.0363\,{\scriptscriptstyle\pm}\,0.0028$ & $0.0363\,{\scriptscriptstyle\pm}\,0.0028$ \\
 & smoother (acc.) & $0.0248\,{\scriptscriptstyle\pm}\,0.0036$ & $0.0539\,{\scriptscriptstyle\pm}\,0.0046$ & $0.0694\,{\scriptscriptstyle\pm}\,0.0037$ \\
 & smoother (res.) & $0.0229\,{\scriptscriptstyle\pm}\,0.0033$ & $0.0556\,{\scriptscriptstyle\pm}\,0.0046$ & $0.0692\,{\scriptscriptstyle\pm}\,0.0040$ \\
 & MaDE & $0.0037\,{\scriptscriptstyle\pm}\,0.0022$ & $0.0000\,{\scriptscriptstyle\pm}\,0.0000$ & $0.0037\,{\scriptscriptstyle\pm}\,0.0022$ \\
\midrule
\multirow{5}{*}{Transformer} & raw & $0.0071\,{\scriptscriptstyle\pm}\,0.0021$ & $0.0251\,{\scriptscriptstyle\pm}\,0.0071$ & $0.0287\,{\scriptscriptstyle\pm}\,0.0077$ \\
 & clamp & $0.0000\,{\scriptscriptstyle\pm}\,0.0000$ & $0.0239\,{\scriptscriptstyle\pm}\,0.0069$ & $0.0239\,{\scriptscriptstyle\pm}\,0.0069$ \\
 & smoother (acc.) & $0.0173\,{\scriptscriptstyle\pm}\,0.0033$ & $0.0282\,{\scriptscriptstyle\pm}\,0.0088$ & $0.0385\,{\scriptscriptstyle\pm}\,0.0088$ \\
 & smoother (res.) & $0.0173\,{\scriptscriptstyle\pm}\,0.0033$ & $0.0282\,{\scriptscriptstyle\pm}\,0.0088$ & $0.0385\,{\scriptscriptstyle\pm}\,0.0088$ \\
 & MaDE & $0.0026\,{\scriptscriptstyle\pm}\,0.0020$ & $0.0000\,{\scriptscriptstyle\pm}\,0.0000$ & $0.0026\,{\scriptscriptstyle\pm}\,0.0020$ \\
\bottomrule
\end{tabular}
}
\end{adjustbox}%
\end{table}

\subsection{Per-Model Breakdown}
\label{app:per-model}

Pooled over all cells on the evaluation windows, the mean Dyn.-K is 0.00718 for MaDE over its 45 cells and 0.17095 for the raw forecasts over their 15 predictor runs. The ratio of MaDE's pooled mean ADE to that of the raw predictors is 1.74.

In all 15 predictor runs, the seed-1 model has the lowest Dyn.-K of the three frozen real-data MaDE models and the seed-0 model the highest. Their means are 0.00230 and 0.01032 respectively, against 0.00891 for the seed-2 model. The seed-1 model has the lowest mean ADE and FDE, 1.0646 m and 2.8443 m, and the seed-0 model the highest, 1.1944 m and 3.1763 m, a factor of 1.122 on ADE. Run by run, the seed-0 model is the highest on both in all 15 runs, and the seed-1 model is below the seed-2 model in 14 of the 15 on ADE and in all 15 on FDE.

Across the 15 predictor runs the raw inequality violation rate spans 0.0228 to 0.0687, a factor of 3.015. After correction it spans 0.00036 to 0.00817, 0.01044 to 0.02135 and 0.00430 to 0.01413 for the models trained with seeds 0, 1 and 2. The seed-0 model has the lowest mean inequality violation rate, 0.00397, and the seed-1 model the highest, 0.01600. Run by run, the seed-0 model is the lowest and the seed-1 model the highest in all 15 runs.

\subsection{Completion-Only Variant}
\label{app:completion-only}

\begin{table}[!t]
\caption{The completion-only variant on the real-data experiment, as the mean $\pm$ the population standard deviation over cells, 5 per predictor family for raw and 15 for completion only and MaDE. Lower is better.}
\label{tab:completion-only}
\centering
\begin{adjustbox}{max width=\linewidth}{\setlength{\tabcolsep}{4pt}
\begin{tabular}{llccc}
\toprule
 & & \multicolumn{2}{c}{Filtered windows} & \multicolumn{1}{c}{All test windows} \\
\cmidrule(lr){3-4} \cmidrule(lr){5-5}
Predictor & Row & Ineq.\ rate (filt.) & Ineq.\ mag.\ (filt.) & Ineq.\ mag.\ (all) \\
\midrule
\multirow{3}{*}{Recurrent} & raw & $0.0572\,{\scriptscriptstyle\pm}\,0.0093$ & $0.0337\,{\scriptscriptstyle\pm}\,0.0048$ & $0.0748\,{\scriptscriptstyle\pm}\,0.0333$ \\
 & completion only & $0.1207\,{\scriptscriptstyle\pm}\,0.0653$ & $0.1023\,{\scriptscriptstyle\pm}\,0.0696$ & $0.3784\,{\scriptscriptstyle\pm}\,0.2393$ \\
 & MaDE & $0.0114\,{\scriptscriptstyle\pm}\,0.0059$ & $0.0034\,{\scriptscriptstyle\pm}\,0.0021$ & $0.0167\,{\scriptscriptstyle\pm}\,0.0107$ \\
\midrule
\multirow{3}{*}{State-space} & raw & $0.0583\,{\scriptscriptstyle\pm}\,0.0057$ & $0.0397\,{\scriptscriptstyle\pm}\,0.0030$ & $0.0897\,{\scriptscriptstyle\pm}\,0.0342$ \\
 & completion only & $0.1295\,{\scriptscriptstyle\pm}\,0.0693$ & $0.1074\,{\scriptscriptstyle\pm}\,0.0709$ & $0.3608\,{\scriptscriptstyle\pm}\,0.1839$ \\
 & MaDE & $0.0111\,{\scriptscriptstyle\pm}\,0.0053$ & $0.0037\,{\scriptscriptstyle\pm}\,0.0022$ & $0.0259\,{\scriptscriptstyle\pm}\,0.0108$ \\
\midrule
\multirow{3}{*}{Transformer} & raw & $0.0358\,{\scriptscriptstyle\pm}\,0.0096$ & $0.0287\,{\scriptscriptstyle\pm}\,0.0077$ & $0.0340\,{\scriptscriptstyle\pm}\,0.0214$ \\
 & completion only & $0.1032\,{\scriptscriptstyle\pm}\,0.0814$ & $0.0877\,{\scriptscriptstyle\pm}\,0.0649$ & $0.4103\,{\scriptscriptstyle\pm}\,0.2573$ \\
 & MaDE & $0.0065\,{\scriptscriptstyle\pm}\,0.0047$ & $0.0026\,{\scriptscriptstyle\pm}\,0.0020$ & $0.0037\,{\scriptscriptstyle\pm}\,0.0062$ \\
\bottomrule
\end{tabular}
}
\end{adjustbox}%
\end{table}

The three rows of Table~\ref{tab:completion-only} come from a single evaluation run and follow the inequality convention of Section~\ref{sec:experiments:metrics}, against the physical constraint set of Section~\ref{sec:experiments:e05}. That convention scores the completion-only variant on the controls it emits. Pooled figures are means over the 15 raw predictor runs and over the same 45 cells for both MaDE rows. On the evaluation windows the pooled violation magnitude is 0.0340 for the raw forecast, 0.0992 for the completion-only variant and 0.0032 for the full operator. Section~\ref{sec:results:ind} gives the rates and the cell-wise comparison of the completion-only variant with its paired raw forecast. On all 54,465 test windows the mean violation magnitude is 0.0662, 0.3832 and 0.0154 in the same order, and the mean violation rate is 0.1377, 0.3207 and 0.0778. In all 45 cells the full operator is below both the completion-only variant and the raw forecast on the violation rate and the magnitude, over the evaluation windows and over all test windows.
\section{Runtime}
\label{app:runtime}

\begin{table}[h]
\caption{Correction cost on the real-data experiment. Lower is better.}
\label{tab:runtime}
\centering
\begin{tabular}{ll}
\toprule
Method & Cost per corrected trajectory \\
\midrule
Raw predictor, amortised in a batch of 16 & $0.0500$ ms \\
Raw predictor, batch size 1 & $0.4739$ ms \\
Clamp, amortised in a batch of 16 & $0.0504$ ms \\
Clamp, batch size 1 & $0.4799$ ms \\
MaDE, amortised in a batch of 16 & $2.5127$ ms \\
MaDE, batch size 1 & $4.9297$ ms \\
\bottomrule
\end{tabular}
\end{table}

Every cost is measured with the timing routine of the evaluation pass, under that pass's own arguments and defaults, at the $0.2$ s step size of the inD evaluation windows. A cell's batch-size-1 figure is the median of five timed calls on one evaluation window, the same window in every cell. Its amortised figure is one timed call on a batch of 16 evaluation windows, divided by 16, and every cell uses the same windows. Each baseline figure is the mean over its 15 cells. The accuracy-tuned and residual-tuned smoothers cost $3.9078$ ms and $3.9205$ ms per trajectory at batch size 1, and $0.2721$ ms and $0.2722$ ms amortised. Every row, MaDE's included, was measured in one run on one device that was idle when the run started, while a second device on the same machine was carrying another process.

Each timed correction covers a 15-step 3 s horizon and runs on one NVIDIA RTX 3090 (24 GB), in a reference implementation in float64 with nothing profiled or fused. The reported costs are scoped to this measured protocol, and no cost is claimed for any other precision or device. Each corrected timestep costs one inverse-model evaluation and one fixed-step Heun solve, followed by the corrector loop. Each corrector iteration costs one gradient of the violation objective with respect to the control plus one Heun re-completion. The evaluation loop early-stops at tolerance $10^{-6}$ and is capped at 50 iterations, so cost tracks the violation actually present in the proposal.

For a fixed MaDE model the evaluation corrector's iteration count depends on the window it corrects. Over the 6,681 evaluation windows and the 45 cells the corrector makes 4,509,675 calls (1,503,225 per MaDE model), where one call corrects one timestep of one window. The pooled count has a median of 0, a 95th percentile of 1, a mean of 0.614 and a maximum of 50 iterations, which is the cap. Of these calls, 0.964\% stop at the cap with the maximum violation still above the tolerance. Each model shares the pooled median, 95th percentile and maximum. For the models trained with seeds 0, 1 and 2 the means are 0.467, 0.850 and 0.525 iterations, and the cap-hit shares are 0.397\%, 1.600\% and 0.896\%.

The iteration counts above pool all evaluation windows, while each batch-size-1 figure times one window. On that window the MaDE cells fall into two cost groups, 38 cells from $1.6224$ ms to $2.8739$ ms and seven from $9.0582$ ms to $35.0782$ ms. Over all 45 cells the batch-size-1 cost has a median of $2.4116$ ms and a mean of $4.9297$ ms. Because every cell times the same window, a spread of that cost across cells cannot come from variation in the data. On that window each MaDE model makes 225 corrector calls, 15 in each of its 15 cells. The models trained with seeds 0, 1 and 2 average 1.964, 2.538 and 0.933 iterations per call there, and stop at the cap on 7, 10 and 2 of their calls. The 19 cap hits fall in 7 of the 45 cells, between 1 and 5 in each, and no call in the other 38 cells runs more than one iteration. Three of the seven are the transformer predictor run with seed 1 under each of the three MaDE models. The other four use the LSTM run with seed 2 or a state-space run with seed 1 or 2. The seven cells with cap hits are the seven slower cells.

At each timestep a batched correction runs as many corrector iterations as the slowest of its trajectories needs, so the amortised figure depends on every window in the batch. Over the 45 cells its mean is $2.5127$ ms, its median $0.6382$ ms, and its range $0.6144$ to $16.0995$ ms. The amortised costs split into a group of 31 cells between $0.6144$ ms and $0.6420$ ms and a group of 14 between $1.7657$ ms and $16.0995$ ms. The batch of 16 includes the window timed at batch size 1, and the group of 14 includes the seven cells that stop at the cap on that window. The iteration counts above are pooled over windows, so they do not show which windows of a batch set its cost.

\paragraph{Deployment envelope.} MaDE at its present cost is scoped to the uses this paper names: vehicle trajectory reconstruction, simulation, batched behavioural forecasting, and offline certification. Real-time onboard use inside a closed-loop control stack is not claimed, and reaching a real-time budget on embedded automotive hardware is left to future work. The amortised cost spreads the cost of one 16-trajectory batch across its members, so it is a throughput figure rather than a latency figure. A stack that corrects one trajectory at a time pays the batch-size-1 cost.

\paragraph{Training compute.} Each training run used one of the machine's two NVIDIA RTX 3090 (24 GB) cards in float64, and the two cards ran separate jobs. Each compute figure is single-GPU wall-clock time taken from recorded run times. On the real-data experiment, MaDE took 47.92 to 71.05 minutes per seed over both training phases, 3.13 GPU-hours over the three seeds. Phase 1 accounts for 0.53 of those hours, a split read from checkpoint times and accurate to about one checkpoint interval. The 15 predictors took 1.34 GPU-hours, at 2.99 to 6.93 minutes per run. The original training of the simulated-system MaDE models, ablations included, has no timing record, so the paper gives no compute figure for it.

\end{document}